\documentclass[10pt,twocolumn,journal]{IEEEtran}

\usepackage{epsfig}
\usepackage{epstopdf}
\usepackage{citesort}
\usepackage{amsmath}
\usepackage{amssymb}
\usepackage{color}
\usepackage{array}
\usepackage{multirow}
\usepackage{algorithm}
\usepackage{algorithmic}
\usepackage{rotating}
\usepackage{graphicx}
\newlength{\figurewidth}
\newlength{\smallfigurewidth}

\begin{document}

\title
{
Single-source Domain Expansion Network for Cross-Scene Hyperspectral Image Classification
}

\author{%
Yuxiang Zhang,~\IEEEmembership{Student Member,~IEEE},
Wei Li,~\IEEEmembership{Senior Member,~IEEE},
Weidong Sun,~\IEEEmembership{Member,~IEEE}, \\
Ran Tao,~\IEEEmembership{Senior Member,~IEEE},
Qian Du,~\IEEEmembership{Fellow,~IEEE}
\thanks{%
	This work was supported by the National Natural Science Foundation
	of China (61922013), partly by Beijing Natural Science Foundation (JQ20021). (corresponding author: Wei Li; liwei089@ieee.org).  }
\thanks{%
Y. Zhang, W. Li and R. Tao are with the School of Information and Electronics, Beijing Institute of Technology, and Beijing Key Laboratory of Fractional Signals and Systems, 100081 Beijing, China (e-mail: zyx829625@163.com, liwei089@ieee.org, rantao@bit.edu.cn).
}
\thanks{%
	Weidong Sun is with the Department of Electronic Engineering, Institute for Ocean Engineering, and the Beijing National Research Center for Information Science and Technology (BNRist), Tsinghua University, Beijing 100084, China (e-mail: wdsun@tsinghua.edu.cn).
}
\thanks{%
	Q. Du is with the Department of Electrical and Computer Engineering, Mississippi State University, Mississippi State, MS 39762 USA.
	(e-mail: du@ece.msstate.edu). }
}

\maketitle
\thispagestyle{empty}
\pagestyle{empty}


\begin{abstract}
Currently, cross-scene hyperspectral image (HSI) classification has drawn increasing attention. It is necessary to train a model only on source domain (SD) and directly transfering the model to target domain (TD), when TD needs to be processed in real time and cannot be reused for training. Based on the idea of domain generalization, a Single-source Domain Expansion Network (SDEnet) is developed to ensure the reliability and effectiveness of domain extension. The method uses generative adversarial learning to train in SD and test in TD. A generator including semantic encoder and morph encoder is designed to generate the extended domain (ED) based on encoder-randomization-decoder architecture, where spatial and spectral randomization are specifically used to generate variable spatial and spectral information, and the morphological knowledge is implicitly applied as domain invariant information during domain expansion. Furthermore, the supervised contrastive learning is employed in the discriminator to learn class-wise domain invariant representation, which drives intra-class samples of SD and ED. Meanwhile, adversarial training is designed to optimize the generator to drive intra-class samples of SD and ED to be separated. Extensive experiments on two public HSI datasets and one additional multispectral image (MSI) dataset demonstrate the superiority of the proposed method when compared with state-of-the-art techniques.
\end{abstract}


\section{Introduction}

With the rapid development of deep learning methods, remote sensing image classification based on Convolutional Neural Network (CNN) has received extensive attention, and achieved excellent performance, particularly in hyperspectral image (HSI) classification \cite{9573256,8105856,9693311}. However, most of CNN-based classification methods need sufficient and accurate labeled samples. In practical applications, the classification performance of new scene data is poor due to the difficulty of collecting labeled samples of remote sensing data and the high cost of manual annotation. It is the most common way to classify the target domain (TD) with a small amount or even no labels by using the source domain (SD) with sufficient labeled samples. However, in practical tasks, the area to be evaluated is often uncertain, and the existing models are limited by the current scene in feature learning. In addition, the acquisition process of HSI is inevitably affected by various factors, such as sensor nonlinearities, seasonal and weather conditions \cite{9127776,8063434}, which lead to variations in spectral reflectance between SD and TD of the same land cover classes. As a result, classification based on CNN has high generalization error and poor interpretation effect in the cross-scene classification task.

Domain adaptation (DA), as a case of transductive transfer learning, reduces domain shift in feature-level and learns domain invariant models. Many methods have been developed for cross-scene classification from the perspective of DA, mainly including related strategies based on statistics, subspace learning, active learning or deep learning. Among them, the maximum mean discrepancy (MMD) criterion \cite{6287330} was the earliest statistical technique used in the cross-scene interpretation. Ganin et al. \cite{ganin2016domain} proposed a Domain Adversarial Neural Network (DANN) for DA, which performs reverse training on generators and discriminators. The training discriminator recognizes the domain, and the training generator tricks the discriminator into learning the invariant feature representation of the domain. Yu et al. developed a Dynamic Adversarial Adaptation Network (DAAN) to solve the problem of dynamic distribution adaptation in an adversarial network \cite{yu2019transfer}. Zhu et al. proposed a Multi-Representation Adaptation Network (MRAN) to accomplish the cross-domain image classification task via multi-representation alignment \cite{zhu2019multi}. In addition, the concept of sub-domains was proposed to improve MMD, and the Deep Subdomain Adaption Network (DSAN) was proposed that used local MDD (LMMD) to align the relevant sub-domains \cite{2020Deepsub}. Class-wise distribution adaptation was designed for HSI cross-scene classification \cite{liu2020class}, and the MMD method based on probability prediction was employed in an adversarial adaptation network to obtain more accurate feature alignment.

The above DA methods have achieved great performance, but the task setting may deviate from actual application. The training samples of DA are labeled SD and unlabeled TD, that is, TD is accessed by the model during training. Domain generalization (DG) is more challenging in task setting than DA, where training samples only include labeled SD. The objective of domain generalization is to learn a model from one or several different but related domains (i.e., diverse training datasets) that generalize well on TD \cite{wang2021generalizing}. In the past few years, domain generalization has made significant progress in computer vision. A Style Normalization and Restitution module (SNR) was proposed to encourage better separation of task-dependent and task-independent features, while ensuring high generalization and high resolution of the network \cite{9157711}. A process of data generation is used to enhance generalization capabilities by increasing the diversity \cite{sicilia2021domain,volpi2018generalizing}. Zhou et al. adversarially trained a transformation network for data augmentation instead of directly updating the inputs by gradient ascent \cite{zhou2020deep}. Li et al. developed a learning framework called Progressive Domain Expansion Network (PDEN) for single domain generalization, which gradually generates multiple domains to simulate various photometric and geometric transformations in TD \cite{li2021progressive}. Domain-adversarial learning is also widely used to learn the domain invariant representation of TD \cite{wang2020unseen,rahman2020correlation}. Peng et al. designed a Deep Adversarial Decoupled Autoencoder (DADA) to decouple class-specific features from class identity \cite{peng2019domain}. 

At present, all the cross-scene HSI classification methods are based on labeled SD data and unlabeled TD data. How to carry out cross-scene classification under the condition that only SD data are available for training has never been considered, which is more challenging. For example, when the limited computing resources of a spaceborne platform make it impossible to re-use TD in training, it is necessary to consider training only according to SD and directly transfering the model to TD. Most of methods do not take into account
the diversity of spectral information, resulting in the lack of effectiveness of the generated samples. Furthermore, multiple augmentation types without considering inter-domain invariance may create unreliable samples that lose discriminant information and are quite distinct from SD. Therefore, while combining spatial-spectral information to generate effective samples, it is necessary to apply template features with domain invariance to ensure their usefulness.

In order to solve the above issues, a DG framework for HSI, called Single-source Domain Expansion Network (SDEnet), is proposed. It covers the domain shift with TD as much as possible, and imposes sufficient reliability constraints and effectiveness constraints on the learning strategy to improve the generalization ability. Specifically, SDEnet includes two components: a generator and a discriminator. In the generator, the sample generated by the single-SD is called the extended domain (ED). To ensure its \textbf{\itshape effectiveness} ({\itshape ED contains specific information of various TDs, not too similar to SD}), a semantic encoder combining spatial-spectral information is designed in SDEnet. In the semantic encoder, the spatial randomization (SpaR) and spectral randomization (SpeR) regarded spatial information and spectral information as the style and content of local patches, respectively, are used for randomization of style representation and content representation. Secondly, in order to ensure its \textbf{\itshape reliability} ({\itshape ED contains domain invariant information, not too different from SD}), a simple morph encoder is designed to extract features with morphological knowledge, which is called template features, so that ED retains SD discriminative information not far away from SD. In the discriminator, the supervised contrastive learning is employed to learn the class-wise domain invariant representation betwee SD, ED and their random linear combination intermediate domain (ID). Furthermore, a supervised contrastive adversarial learning strategy is designed to improve the expansion ability of the generator.

The main contributions of this work are summarized as follows.

\begin{itemize}
	\item To the best of our knowledge, this is the first work to propose a DG framework for cross-scene HSI classification, which shows that DG has more practical application significance than the traditional DA.
	
	\item The semantic encoder combining spatial-spectral information is more suitable for HSI, which carries out spatial-level and spectral-level randomization to ensure the effectiveness of generated samples.
	
	\item Morph encoder is designed to extract template features with domain invariance, and collaborates with semantic encoder for domain extension to ensure the reliability of generated samples.
	
	\item The supervised contrastive adversarial learning strategy is developed to improve the generalization capability, where the generator and discriminator compete by iteratively generating out-of-domain data and learning class-wise domain invariant representation.
\end{itemize}

The rest of the paper is organized as follows. Section II introduces relevant concepts of DG and contrastive learning. Section III elaborates on the proposed SDEnet. The extensive experiments and analyses are presented in Section IV. Finally, conclusions are drawn in Section V.

\section{Related Work}
\label{sec:related-works}

\begin{table}[]
	\begin{minipage}[!t]{\columnwidth}
	\renewcommand{\arraystretch}{1.3}
	\caption{\label{table:DADG}
	Comparison between domain adaptation and domain generalization}
			\begin{tabular}{cccc}
				\hline
				Learning paradigm                                                              & Training data                              & Test data        & Test access       \\ \hline
				Domain adaptation                                                              & ${\cal S}^{src}, {\cal S}^{tar}$           & ${\cal S}^{tar}$ & $\surd$           \\
				\begin{tabular}[c]{@{}c@{}}Single-source \\ Domain generalization\end{tabular} & ${\cal S}^{src}$                             & ${\cal S}^{tar}$   & ${\rm{ \times }}$ \\
				\begin{tabular}[c]{@{}c@{}}Multi-source \\ Domain generalization\end{tabular}  & ${\cal S}^{1}, {\cal S}^{2}... {\cal S}^{n}$ & ${\cal S}^{n+1}$ & ${\rm{ \times }}$ \\ \hline
			\end{tabular}
\end{minipage}
\end{table}

\subsection{Domain Generalization (DG)}

DG is more challenging than DA, because DG aims to learn the model through SD data and does not need to access TD in the training phase. The model can be extended to TD in the inference stage. A comparison between DA and DG is listed in Table \ref{table:DADG}, where the ${\cal S}$ represents domain. The existing DG methods can be divided into two categories: learning the domain invariant representation and data manipulation.

The key idea of the first category is to reduce the domain shift between multiple SD domain representations, which is mainly applied to multi-source DG. The most typical strategy is the explicit feature alignment. Some methods explicitly minimize the feature distribution divergence by minimizing MMD \cite{wang2020transfer}, second-order correlation \cite{sun2016deep}, Wasserstein distance \cite{zhou2020domain} of domains. Domain adversarial learning is widely used to learn domain invariant representation. Most methods train discriminators to distinguish domains, while training generators intend to fool discriminators to learn domain invariant representation. The most typical method is Domain Adversarial Neural Network (DANN) \cite{ganin2016domain} proposed by Ganin et al.

Data manipulation is mainly applied to single-source DG. Such methods generally augment or generate out-of-domain samples related to SD, and then use these samples to train the model with the SD, and transfer to TD. Data augmentation is mainly based on the augmentation, randomization, and transformation of input data, and improves the generalization performance of the model by reducing overfitting. Typical augmentation operations include  flipping, rotation, scaling, cropping, adding noise, etc. Data generation creates diversified and abundant data to help generalization. For example,  Variational Auto-encoder (VAE) \cite{kingma2013auto} and Generative Adversarial Networks (GAN) \cite{goodfellow2014generative} are often used for these purposes. In addition, Adaptive Instance Normalization (AdaIN) \cite{karras2019style}, Mixup \cite{zhang2017mixup} and other strategies are also used. AdaIN is employed in the proposed method to achieve randomization. The AdaIN operation is defined as,
\begin{figure}[tp] \small
	\begin{center}
		\centering
		\epsfig{width=0.6\figurewidth,file=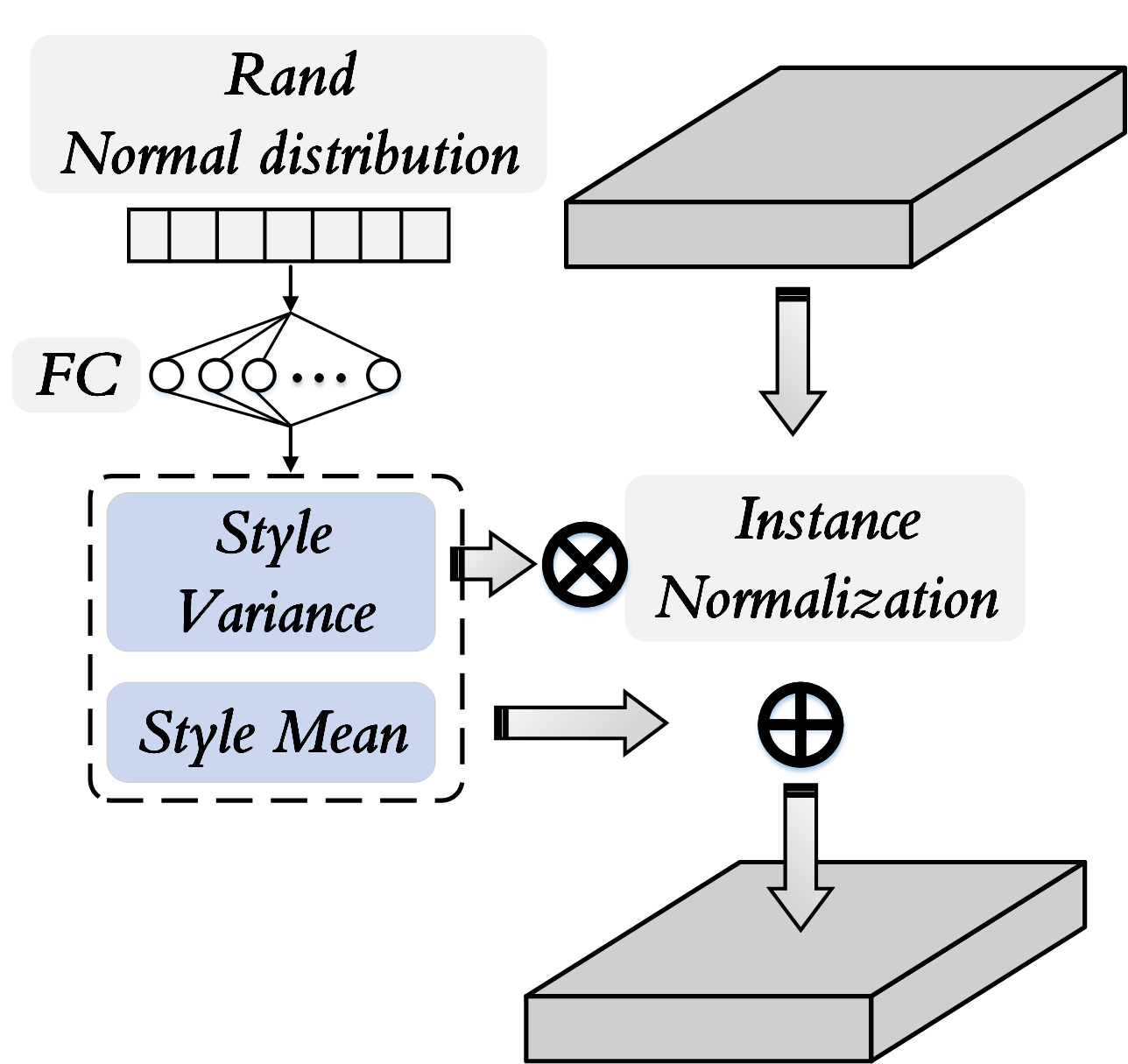}
		\caption{\label{fig:AdaIN}
			The AdaIN calculation process for data generation. The random noise is mapped to the style mean and style variance through the full connection layer, and then applied to the normalized feature map.}
	\end{center}
	\vspace{-2em}
\end{figure}

\begin{equation}\label{eq:1}
AdaIN\left( {\bf{z},\bf{n}} \right) = F{C_1}\left( \bf{n} \right)\frac{{\bf{z} - \mu (\bf{z})}}{{\sigma (\bf{z})}} + F{C_2}(\bf{n})
\end{equation}
where $\bf{z}$ is the normalized feature map, $\bf{n}$ is random noise  $\bf{n} \sim N\left( {0,1} \right)$ and the fully connected layer is denoted as $FC$. The AdaIN calculation process is shown in Fig.~\ref{fig:AdaIN}. 

\subsection{Contrastive Learning}

Contrastive learning is a popular self-supervised pre-training method for image classification in recent years. The core idea is to train a model by automatically constructing similar positive sample pairs and dissimilar negative sample pairs, so that positive pairs are closer in the projection space, while negative pairs are far away. For any sample $\bf{x}$, contrastive methods aim to learn an feature extractor $F$ such that:
\begin{equation}\label{eq:2}
S\left( {F\left( \bf{x} \right),F\left( {{\bf{x}^ + }} \right)} \right) \gg S\left( {F\left( \bf{x} \right),F\left( {{\bf{x}^ - }} \right)} \right)
\end{equation}
where $\bf{x}^+$ is a sample similar to $\bf{x}$, referred to as a positive sample, $\bf{x}^-$ is a sample dissimilar to $\bf{x}$, referred to as a negative sample, $S(\bullet)$ function is a metric that measures the similarity between two features, and $\bf{x}$ is commonly referred to as an ``anchor'' sample. To optimize for this property, the InfoNCE loss  \cite{oord2018representation} is generally constructed to correctly classify positive samples and negative samples,
\begin{equation}\label{eq:3}
{{\cal L}_{NCE}} =  - \log \frac{{\exp \left( {{{S\left( {F\left( {\bf{x}} \right),F\left( {{{\bf{x}}^ + }} \right)} \right)} \mathord{\left/
					{\vphantom {{S\left( {F\left( {\bf{x}} \right),F\left( {{{\bf{x}}^ + }} \right)} \right)} \tau }} \right.
					\kern-\nulldelimiterspace} \tau }} \right)}}{{\sum\nolimits_{j = 1}^{N - 1} {\exp \left( {{{S\left( {F\left( {\bf{x}} \right),F\left( {{{\bf{x}}^ - }} \right)} \right)} \mathord{\left/
						{\vphantom {{S\left( {F\left( {\bf{x}} \right),F\left( {{{\bf{x}}^ - }} \right)} \right)} \tau }} \right.
						\kern-\nulldelimiterspace} \tau }} \right)} }}
\end{equation}
where $S(\bullet)$ generally uses dot product or cosine distance, and $\tau$ is a temperature hyper-parameter that controls the sensitivity of $S(\bullet)$. Currently, many experiments indicate that $\tau$ should set a relatively small value, generally set to 0.1 or 0.2 \cite{wu2018unsupervised,chen2020simple}. The sum in the denominator is calculated over one positive and $N-1$ negative pairs in the same minibatch. The InfoNCE loss should encourage the $S(\bullet)$ function to assign large values to positive samples and small values to negative samples.

Existing contrastive learning methods have various strategies to generate positive and negative samples. MoCo \cite{he2020momentum} maintains the running momentum encoder and a finite queue of previous samples. Tian et al. \cite{tian2020contrastive} consider all multi-view samples produced by the minibatch method, while SimCLR \cite{chen2020simple} uses the momentum encoder and all generated samples within the minibatch.

\begin{table}[!t]
	\begin{minipage}[!t]{\columnwidth}
		\renewcommand{\arraystretch}{1.3}
		\caption{\label{table:Abbreviation}
			Summary of abbreviations.}
		\centering
		\setlength{\tabcolsep}{0.6mm}{
			\begin{tabular}{ll}
				\hline
				\multicolumn{1}{c}{Abbreviation}                      & Description                                                 \\ \hline
				HSI                                                   & Hyperspectral image                                     \\
				MSI                                                   & Multispectral image                                     \\
				SD                                                    & Source domain                                             \\
				TD                                                    & Target domain                                             \\
				ED                                                    & Extended domain                                            \\
				ID                                                    &  Intermediate domain                                      \\
				DA                                                    & Domain adaptation                                       \\
				DG                                                   & Domain generalization                             \\
				SpaR                                                   & Spatial randomization                             \\
				SpeR                                                   & Spectral randomization          \\ \hline
		\end{tabular}}
	\end{minipage}
	\\[12pt]
	\begin{minipage}[!t]{\columnwidth}
		\renewcommand{\arraystretch}{1.3}
		\caption{\label{table:Notations}
			Notations of variables.}
		\centering
		\setlength{\tabcolsep}{1mm}{
			\begin{tabular}{ll}
				\hline
				Notations                               & Description                                                 \\ \hline
				${{\bf{X}}}$, ${\bf{\hat X}}$ and ${\bf{\tilde X}}$       & Source, extened and intermediate domain                 \\
				${{{\bf{z}}_{spa}}}$                              & Spatial feature map                      \\
				${{{\bf{z}}_{spe}}}$       & Spectral embedding feature\\
				$G$       & Generator   \\
				$D$       & Discriminator   \\
				${f_{emb}}$       & Feature extractor    \\
				$C$       & Classification head  \\
				$P$       & Projection head \\ \hline
				&                                                             \\
				&
		\end{tabular}}
	\end{minipage}
\vspace{-3em}
\end{table}

\section{Proposed Single-source Domain Expansion Network}
\label{sec:proposed}

\begin{figure*}[tp] \small
	\vspace{-2em}
	\begin{center}
		\centering
		\epsfig{width=2.2\figurewidth,file=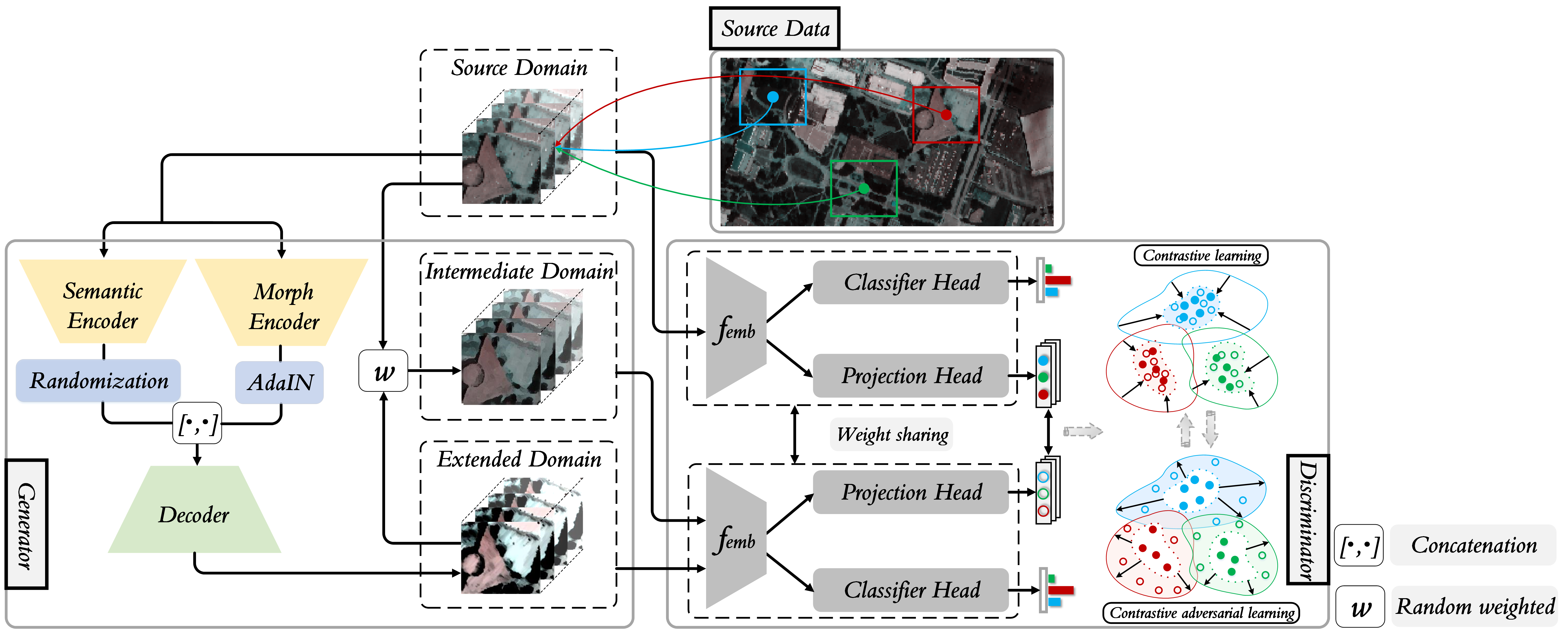}
		\caption{\label{fig:Flowchart}
			Flowchart of the proposed SDEnet, including generator composed of semantic encoder and morph encoder, and discriminator using multiple domain learning class-wise domain invariant representation. Red, green and blue represent three classes respectively. The embedding features of SD output by the projection head are shown as solid circles, while the embedding features of ID and ED are shown as hollow circles.}
	\end{center}
\vspace{-2em}
\end{figure*}

Notations used in this paper is summarized in Table \ref{table:Notations}. Assume that ${{\bf{X}}} = \left\{ {{\bf{x}}_i} \right\}_{i = 1}^{{N}}\in\mathbb{R}{^d}$ is the data from SD, and ${{\bf{Y}}}= \left\{ {{\bf{y}}_i} \right\}_{i = 1}^{{N}}$ is the corresponding class labels. Here, $d$ and $N$ denote the dimension of data and the number of source samples, respectively. The proposed SDEnet includes a generator and a discriminator, as shown in Fig.~\ref{fig:Flowchart}. The sample of 13$\times $13$\times d$ spatial patch in HSI is selected from SD and sent to the generator for semantic encoding and morphological encoding respectively. The semantic encoder uses the  1$\times$1 convolution kernel and the convolution kernel with the same size as patch to form spatial and spectral features. After performing spatial and spectral randomization, the deconvolution is used to map back to the feature map of patch size. In addition, a simple morph encoder is constructed by using the Dilation2D convolution and Erosion2D convolution, and the template features with domain invariance are extracted and randomized by AdaIN. The output of two encoders are concatenated and input into the decoder to generate ED. SD and ED are randomly linear weighted to obtain ID. Then, SD, ID and ED are used as input of discriminator to pass through feature extractor ${f_{emb}}$ with shared weight, the classification head $C$ is used to calculate the cross entropy loss, and the projection head $P$ outputs the embedding features to construct positive and negative pairs for contrastive and adversarial learning.

The proposed method ensures the effectiveness and reliability of ED from two aspects: structure and loss function. \textbf{\itshape Effectiveness}: (1) structure, spatial-spectral generation strategy (spatial randomization and spectral randomization); (2) loss, adversarial training with supervised contrastive learning. \textbf{\itshape Reliability}: (1) structure, morphological knowledge as domain invariance feature (template feature); (2) loss, classification loss of ED and ID.

\subsection{ Domain Expansion Generator}

For the single-source DG task, a generator $G$ is designed in SDEnet to generate the ED ${\bf{\hat X}}$ with a certain domain shift from single SD ${\bf{X}}$. The random linear combination $w$ is applied to ${\bf{X}}$ and ${\bf{\hat X}}$ to compute ID,

\begin{equation}\label{eq:4}
{\bf{\tilde X}} = w{\bf{X}} + \left( {1 - w} \right){\bf{\hat X}}
\end{equation}
where ${\bf{\tilde X}}$ represents ID. When SD and ED have large domain shift, the ID can be used as their transition to alleviate the learning pressure of model and ensure that the domain invariant features is learned. In order to make full use of HSI data characteristics and update an effective and reliable ${\bf{\hat X}}$, semantic encoder and morph encoder are designed respectively. Following the common practice \cite{karras2019style,nam2018batch,lee2019srm}, we utilize the channel-wise mean and standard deviation of embedding features as style representation, and the difference is that spectral information is used as content representation.

\subsubsection{\textbf{Semantic Encoder}}

\begin{figure}[tp] \small
	\begin{center}
		\centering
		\epsfig{width=1.10\figurewidth,file=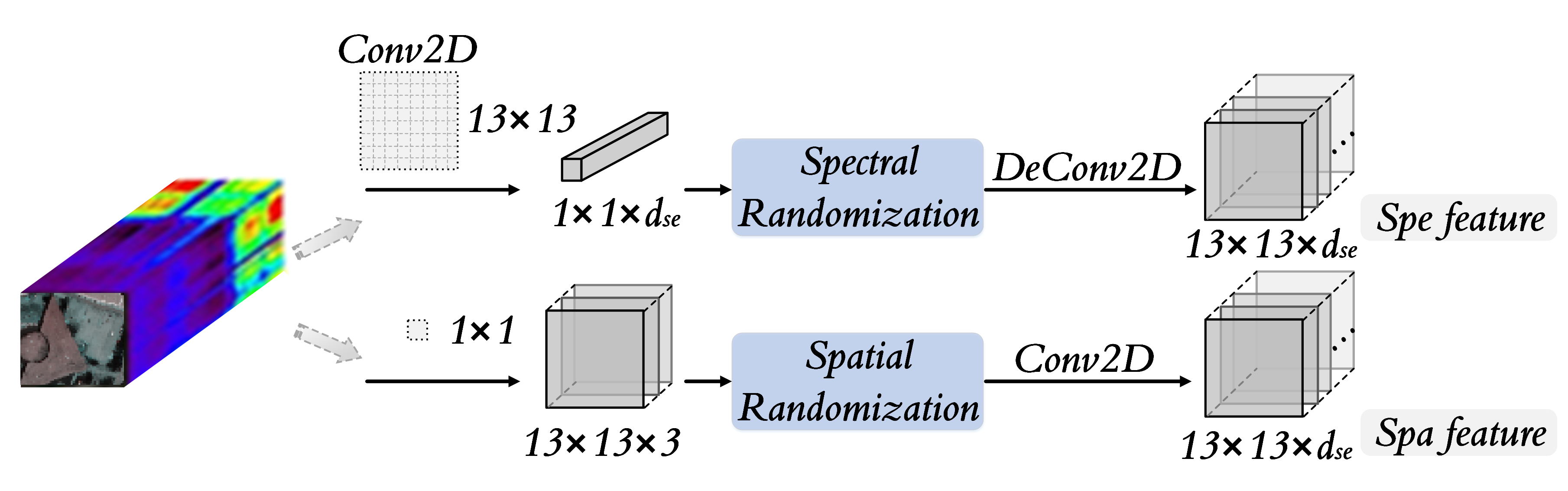}
		\caption{\label{fig:semantic}
			The flowchart of semantic encoder consisting of spatial randomization and spectral randomization.}
	\end{center}
\vspace{-2em}
\end{figure}

Considering that HSI is a data collection with strong spatial recognition and multi-band spectral information, the generation flow of spatial dimension and spectral dimension is carried out in the semantic encoder, as shown in Fig.~\ref{fig:semantic}, to realize the generation strategy of spatial dimension replacement style and spectral dimension replacement content.

In the spatial generation flow, the 1$\times$1 convolution kernel is used to reduce the dimension of a spatial patch to 13$\times$13$\times$3 and treat it as a feature map ${{{\bf{z}}_{spa}}}$ representing spatial information. The style representations of the feature maps in minibatch are then calculated, that is the channel-wise mean $\mu ({{\bf{z}}_{spa}})$ and standard deviation $\sigma ({{\bf{z}}_{spa}})$,
\begin{equation}\label{eq:5}
{\mu ({{\bf{z}}_{spa}}) = \frac{1}{{HW}}\sum\limits_{h = 1}^H {\sum\limits_{w = 1}^W {{{\bf{z}}_{spa}}} } }
\end{equation}

\begin{equation}\label{eq:6}
{\sigma ({{\bf{z}}_{spa}}) = \sqrt {\frac{1}{{HW}}\sum\limits_{h = 1}^H {\sum\limits_{w = 1}^W {{{\left( {{{\bf{z}}_{spa}} - \mu ({{\bf{z}}_{spa}})} \right)}^2}} } } }
\end{equation}
where $H$ and $W$ denote length and width of spatial patch, respectively. The following operation is performed in SpaR: $\mu ({{\bf{z'}}_{spa}})$, $\sigma ({{\bf{z'}}_{spa}})$ corresponding to ${{\bf{z'}}_{spa}}$ are randomly selected, and the adaptive linear style combination with $\mu ({{\bf{z}}_{spa}})$ and $\sigma ({{\bf{z}}_{spa}})$ is carried out to obtain $\hat \mu$ and $\hat \sigma$,
\begin{equation}\label{eq:7}
\begin{array}{l}
\hat \mu  = \alpha \mu ({{\bf{z}}_{spa}}) + (1 - \alpha )\mu ({{{\bf{z'}}}_{spa}})\\
\hat \sigma  = \alpha \sigma ({{\bf{z}}_{spa}}) + (1 - \alpha )\sigma ({{{\bf{z'}}}_{spa}})
\end{array}
\end{equation}
where $\alpha$ is an adaptive learning parameter. Then the noise of AdaIN in Eq. \ref{eq:1} is replaced with $\hat \mu$ and $\hat \sigma$ as,
\begin{equation}\label{eq:8}
SpaR\left( {{{\bf{z}}_{spa}},{{{\bf{z'}}}_{spa}}} \right) = \hat \mu \frac{{{{\bf{z}}_{spa}} - \mu ({{\bf{z}}_{spa}})}}{{\sigma ({{\bf{z}}_{spa}})}} + \hat \sigma
\end{equation}
In SpaR, the contents of ${{\bf{z}}_{spa}}$ are kept and replaced with style $\hat \mu$ and $\hat \sigma$. Then, the Spa feature $SpaR\left( {{{\bf{z}}_{spa}},{{{\bf{z'}}}_{spa}}} \right)$ is mapped back to a 13$\times $13$\times d_{se}$ feature map.

Different from natural images, the spatial configuration is regarded as the content representation of image. In HSI, the spectral information in a spatial patch represents the class characteristics. Therefore, in the spectral generation flow, the 13$\times $13 convolution kernel is used to compress spatial information to the spectral embedding features ${{{\bf{z}}_{spe}}}$ of 1$\times $1$\times d_{se}$ size, and send them into SpeR for random content replacement in a minibatch. Here the AdaIN method is used to maintain the style of ${{{\bf{z}}_{spe}}}$ and replace its content with randomly selected ${{{{\bf{z'}}}_{spe}}}$ from minibatch. The mean and standard deviation of ${{{\bf{z}}_{spe}}}$ is regarded as the noise in Eq. \ref{eq:1},
\begin{equation}\label{eq:9}
SpeR\left( {{{\bf{z}}_{spe}},{{{\bf{z'}}}_{spe}}} \right) = \sigma ({{\bf{z}}_{spe}})\frac{{{{{\bf{z'}}}_{spe}} - {\bf{\mu }}({{{\bf{z'}}}_{spe}})}}{{\sigma ({{{\bf{z'}}}_{spe}})}} + {\bf{\mu }}({{\bf{z}}_{spe}})
\end{equation}
Then the Spe feature $SpeR\left( {{\bf{z}_{spe}},{\bf{z'}_{spe}}} \right)$ is mapped back to the 13$\times $13$\times d_{se}$ feature map by deconvolution 2D operation.

\subsubsection{\textbf{Morph Encoder}}

\begin{figure}[tp] \small
	\begin{center}
		\centering
		\epsfig{width=1\figurewidth,file=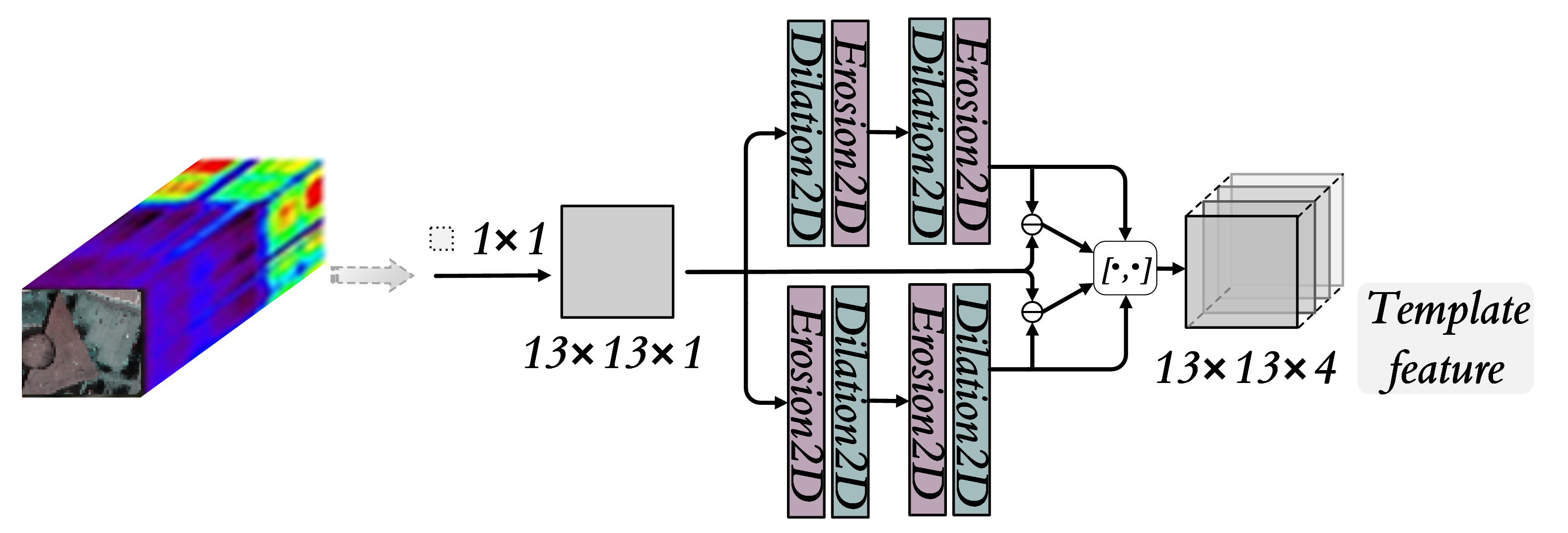}
		\caption{\label{fig:morph}
			The flowchart of morph encoder consisting of Dilation2D and Erosion2D.}
	\end{center}
\vspace{-2em}
\end{figure}

The semantic encoder has completed the generation of various samples, ensuring the effectiveness of ED from the spatial dimension and spectral dimension. However, it is inevitable to generate invalid samples with strong noise in the process of randomization. This kind of samples without domain invariance and SD discriminative information may bring negative transfer effects. It is well known that morphological structure elements are used to discover structures in images. In order to automatically learn the hierarchy of features from data, morphological operations (erosion and dilation, etc.) are arranged in the form of a network and the structural elements are automatically learned \cite{2019morphological}. Furthermore, the spatial patches extracted from HSI have sufficient structural information. In the cross-scene classification, the gap is mainly reflected in the spectral dimension shift, while the structural information of the spatial dimension can be regarded as the domain invariant representation. Therefore, a morph encoder is designed to ensure the reliability of ED in the SDEnet, as shown in Fig.~\ref{fig:morph}.

The gray or color image can be effectively processed by using a two-dimensional morphological operator \cite{2020image}. Therefore, the 13$\times $13$\times d$ spatial patch is reduced to 13$\times$13$\times$1, ${{\bf{z}}_m}$, which is analogous to a grayscale image. Then it is sent into the morphological network to extract the template feature, where the most basic morphological operation dilation and erosion are used to form the morphological network. Dilation2D convolution (or Erosion2D convolution) calculates the maximum  (minimum) of point-to-point sum (difference) between ${{\bf{z}}_m}$ local values and structural elements (kernels),
\begin{equation}\label{eq:10}
\!\!\!\begin{array}{*{20}{l}}
{Dilation2D\left( {{{\bf{z}}_m},{{\bf{w}}_d}} \right)(p,q)\!\!=\!\!\mathop {\max } \left( {{{\bf{z}}_m}(p\!+\!i,q\!+\!j)\!\!+\!\!{{\bf{w}}_d}(i,j)} \right)}\\
{Erosion2D\left( {{{\bf{z}}_m},{{\bf{w}}_e}} \right)(p,q)\!\!=\!\! \mathop {\min }\left( {{{\bf{z}}_m}(p\!+\!i,q\!+\!j)\!-\!{{\bf{w}}_e}(i,j)} \right)}
\end{array}
\end{equation}
where ${{{\bf{w}}_d}}$ and ${{{\bf{w}}_e}}$ are the dilation and erosion structural element, respectively, $p$ and $q$ represent the location of the central pixel of the current local patch. In addition, the inputs are padded with zero.

Some complex morphological operators based on Dilation2D and Erosion2D are constructed in morph encoder. As shown in Fig.~\ref{fig:morph}, the upper branch is the opening operation (Dilation2D to Erosion2D), and the lower branch is the closing operation (Erosion2D to Dilation2D). These two branches carry out twice opening and closing operations without sharing weights. Further, the top hat (${{\bf{z}}_m}-$ opening) and black hat (closing$-{{\bf{z}}_m} $) are built by residual connection with ${{\bf{z}}_m}$. Finally, the four output morphological feature maps are concatenated to form the template features, and AdaIN is used for randomization.

\subsection{Domain Invariant Discriminator}

The discriminator $D$ consists of a feature extractor ${f_{emb}}$, a classification head $C$ and a projection head $P$, which receives SD and generator output ED and ID for learning class-wise domain invariant representation. ${f_{emb}}$ is composed of Conv2d-ReLU-MaxPool2d blocks stacked twice and domain features is output by two layers of FC. $C$ focuses on classification tasks, outputs prediction probabilities and calculates cross-entropy loss,
\begin{equation}\label{eq:11}
{{\cal L}_{ce}}\left( {{{\bf{p}}_i},{{\bf{y}}_i}} \right) =  - \sum\limits_c {y_i^c\log p_i^c} 
\end{equation}
where ${\bf{y}}_i$ is the one-hot encoding of the label information of ${{\bf{x}}_i}$, $c$ is the index of class, and ${{\bf{p}}_i}$ is the predicted probability output obtained by $C$. Therefore, the classification loss for SD is defined as,
\begin{equation}\label{eq:12}
{{{\cal L}_{SD}}\left( {{\bf{X}},{\bf{Y}}} \right) = \frac{1}{N}\sum\limits_i {{{\cal L}_{ce}}\left( {C\left( {{{\bf{x}}_i}} \right),{{\bf{y}}_i}} \right).} }
\end{equation}

In order to ensure the reliability of ED, $D$ is required to correctly predict it. In addition, ID as a transition between SD and ED which also needs to be correctly predicted to reduce the pressure on $D$ to learn ED domain features. ED and ID have the same label space as SD, so the label ${\bf{Y}}$ of SD is used to calculate the classification loss of ED and ID,

\begin{equation}\label{eq:13}
{{\cal L}_{ED}}( {{\bf{\hat X}},{\bf{Y}}}) = \frac{1}{N}\sum\limits_i {{{\cal L}_{ce}}\left( {C\left( {{{{\bf{\hat x}}}_i}} \right),{{\bf{y}}_i}} \right)}
\end{equation}

\begin{equation}\label{eq:14}
{{\cal L}_{ID}}( {{\bf{\tilde X}},{\bf{Y}}}) = \frac{1}{N}\sum\limits_i {{{\cal L}_{ce}}\left( {C\left( {{{{\bf{\tilde x}}}_i}} \right),{{\bf{y}}_i}} \right).} 
\end{equation}\\

In SDEnet, $P$ is only a layer of FC, and it outputs embedding features for supervised contrastive adversarial learning. Firstly, a supervised contrastive learning is introduced to encourage $D$ to learn class-wise domain invariant representation,
\begin{equation}\label{eq:15}
{{\cal L}_{supcon}} =  - \sum\limits_{i = 0}^N {\frac{1}{{|P(i)|}}} \sum\limits_{p \in P(i)} {\log } \frac{{\exp \left( {{{S\left( {{{\bf{z}}_i},{\bf{z}}_p^ + } \right)} \mathord{\left/
					{\vphantom {{S\left( {{{\bf{z}}_i},{\bf{z}}_p^ + } \right)} \tau }} \right.
					\kern-\nulldelimiterspace} \tau }} \right)}}{{\sum\limits_{a \in A(i)} {\exp \left( {{{S\left( {{{\bf{z}}_i},{\bf{z}}_a^ - } \right)} \mathord{\left/
						{\vphantom {{S\left( {{{\bf{z}}_i},{\bf{z}}_a^ - } \right)} \tau }} \right.
						\kern-\nulldelimiterspace} \tau }} \right)} }}
\end{equation}
where for each embedding feature ${{\bf{z}}_i}$ in minibatch, ${P(i)}$ and ${A(i)}$ are the positive and negative sample sets, ${|P(i)|}$ is the number of positive samples, ${{\bf{z}}_p^ + }$ and ${{\bf{z}}_a^ - }$ are one of the positive and negative samples. In the process of optimizing $D$, the features belonging to the same class in SD, ED and ID are put into ${P(i)}$, and the features outside the class are put into ${A(i)}$. $D$ is optimized by ${{\cal L}_{supcon}}$ to make the samples belonging to the same class closer and the samples belonging to different classes farther (e.g., the contrast learning in Fig.~\ref{fig:Flowchart}, the solid circle and hollow circle of the same colors are compact, and different colors are separated). This enables $D$ to learn class-wise shared representations from samples of the same class.

Considering the effectiveness of $G$ to generate ED, adversarial learning is designed to optimize $G$, which is opposite to the direction of $D$ optimization. It is also based on the supervised contrastive learning loss. The difference is that samples of the $c$-th class in SD are taken as positive samples, and samples of the $c$-th class in ED or ID are taken as negative samples,
\begin{equation}\label{eq:16}
{{\cal L}_{adv}} =  - \sum\limits_c {\sum\limits_{i = 0}^{{n_c}} {\frac{1}{{|{P_c}(i)|}}} \sum\limits_{p \in {P_c}(i)} {\log } \frac{{\exp \left( {{{S\left( {{{\bf{z}}_i},{\bf{z}}_p^ + } \right)} \mathord{\left/
						{\vphantom {{S\left( {{{\bf{z}}_i},{\bf{z}}_p^ + } \right)} \tau }} \right.
						\kern-\nulldelimiterspace} \tau }} \right)}}{{\sum\limits_{a \in {A_c}(i)} {\exp \left( {{{S\left( {{{\bf{z}}_i},{\bf{z}}_a^ - } \right)} \mathord{\left/
							{\vphantom {{S\left( {{{\bf{z}}_i},{\bf{z}}_a^ - } \right)} \tau }} \right.
							\kern-\nulldelimiterspace} \tau }} \right)} }}}
\end{equation}
where $n_c$ is the number of samples of the $c$-th class, ${{P_c}(i)}$ and ${{A_c}(i)}$ are the positive and negative sample sets of the $c$-th class, respectively. Here, ED or ID is randomly selected to construct the negative sample set in each iteration. In addition, the supervision labels corresponding to ${{P_c}(i)}$ and ${{A_c}(i)}$ of the $c$-th class are set to 0 and 1. In the process of optimizing $G$, the samples belonging to the $c$-th class in SD and the samples belonging to the $c$-th class in ED or ID are separated from each other (e.g., the contrastive adversarial learning in Fig.~\ref{fig:Flowchart}, the solid circles and hollow circles with the same color are separated). Through adversarial training, $G$ generate ED from which $D$ cannot extract domain invariant representations. It indicates that ED contains domain-specific information.

\subsection{Training Phase}

In SDEnet, $G$ and $D$ are optimized separately. Firstly, $D$ is optimized, including ${f_{emb}}( \cdot ;{{\bf{\theta }}_f})$, $C( \cdot ;{{\bf{\theta }}_C})$ and $P( \cdot ;{{\bf{\theta }}_P})$, with SD and generated ED and ID as inputs. The optimization objective of $D$ is the weight combination of Eqs. \ref{eq:12}-\ref{eq:15} as follows,
\begin{equation}\label{eq:17}
\mathop {\min }\limits_{{{\bf{\theta }}_f},{{\bf{\theta }}_C},{{\bf{\theta }}_P}} {\cal L} = {{\cal L}_{SD}} + {{\cal L}_{ED}} + {{\cal L}_{ID}} + \lambda_1 {{\cal L}_{supcon}}
\end{equation}
where $\lambda$ is a hyper-parameter for balancing the supervised contrastive learning loss. To optimize ${G}( \cdot ;{{\bf{\theta }}_G})$, in addition to solving Eq. \ref{eq:15}, Eq. \ref{eq:13} is used as an auxiliary loss to ensure ED in the correct label space,
\begin{equation}\label{eq:18}
\mathop {\min }\limits_{{{\bf{\theta }}_G}} {\cal L} = {{\cal L}_{ED}} + \lambda_2 {{\cal L}_{adv}}
\end{equation}
where $\lambda$ is a hyper-parameter for balancing the adversarial learning loss. For simplicity, $\lambda_1$ in Eq. \ref{eq:17} and $\lambda_2$ in Eq. \ref{eq:18} are set to the same value.

\begin{figure*}[tp] \small
	\vspace{-2em}
	\begin{center}
		\setlength\tabcolsep{6pt}
		\begin{tabular}{cc}
			\epsfig{width=1\figurewidth,file=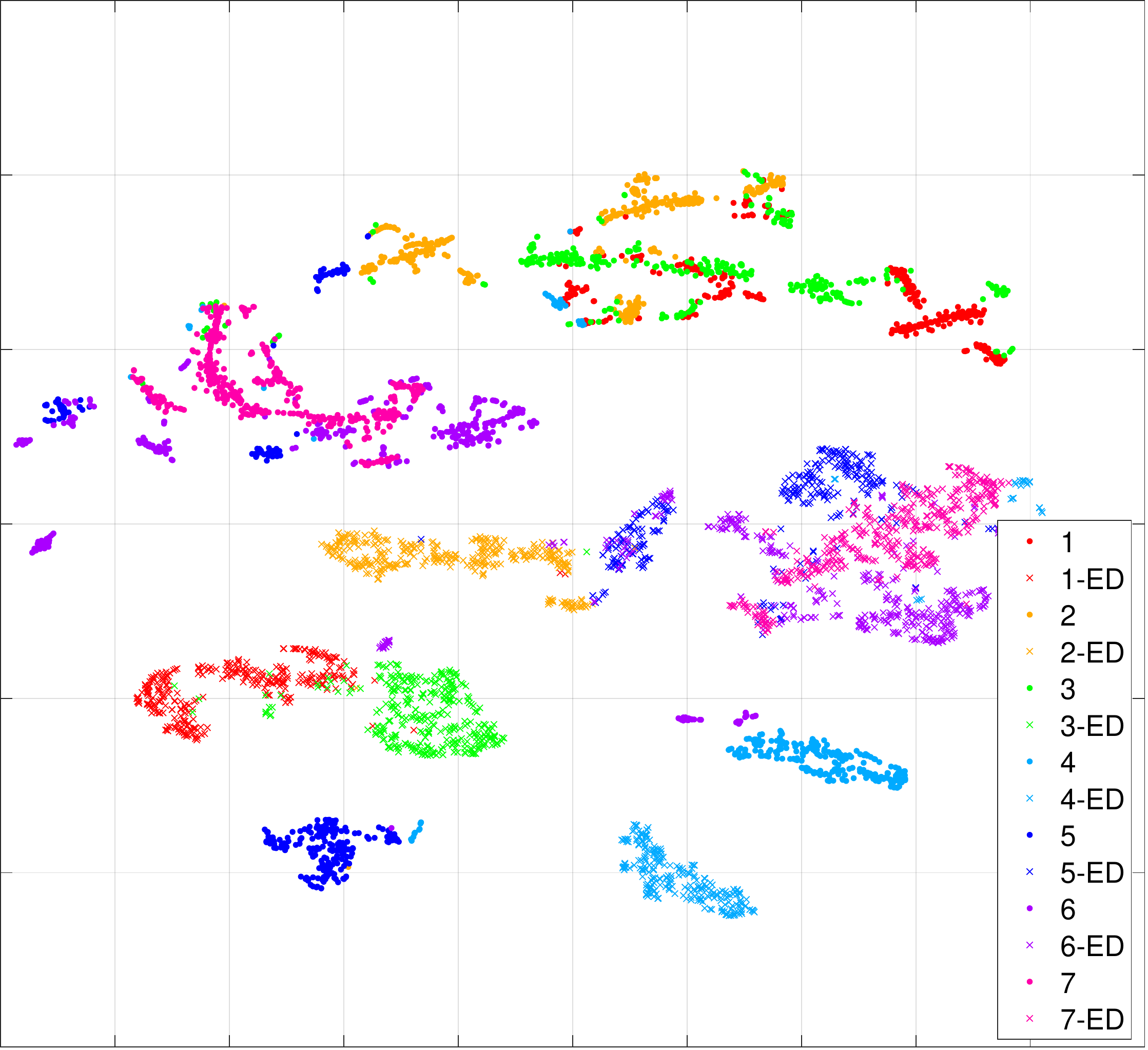} &
			\epsfig{width=1\figurewidth,file=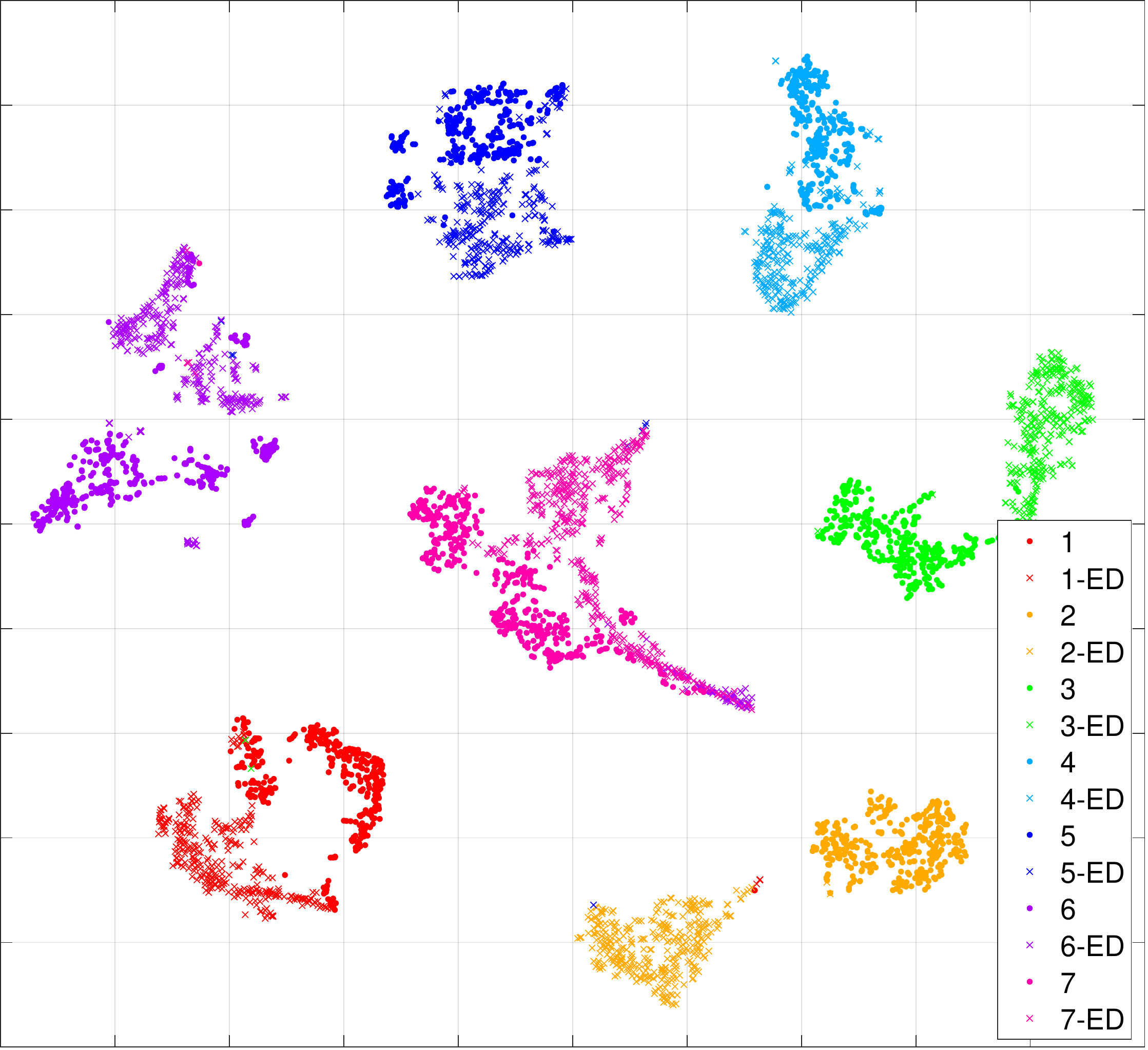}  \\
			(a) Original samples from SD, and ED output by $G$  & (b) SD features and ED features output by $P$ \\  [0.5em]
		\end{tabular}
		\setlength\tabcolsep{6pt}
		\begin{tabular}{cc}
			\epsfig{width=1\figurewidth,file=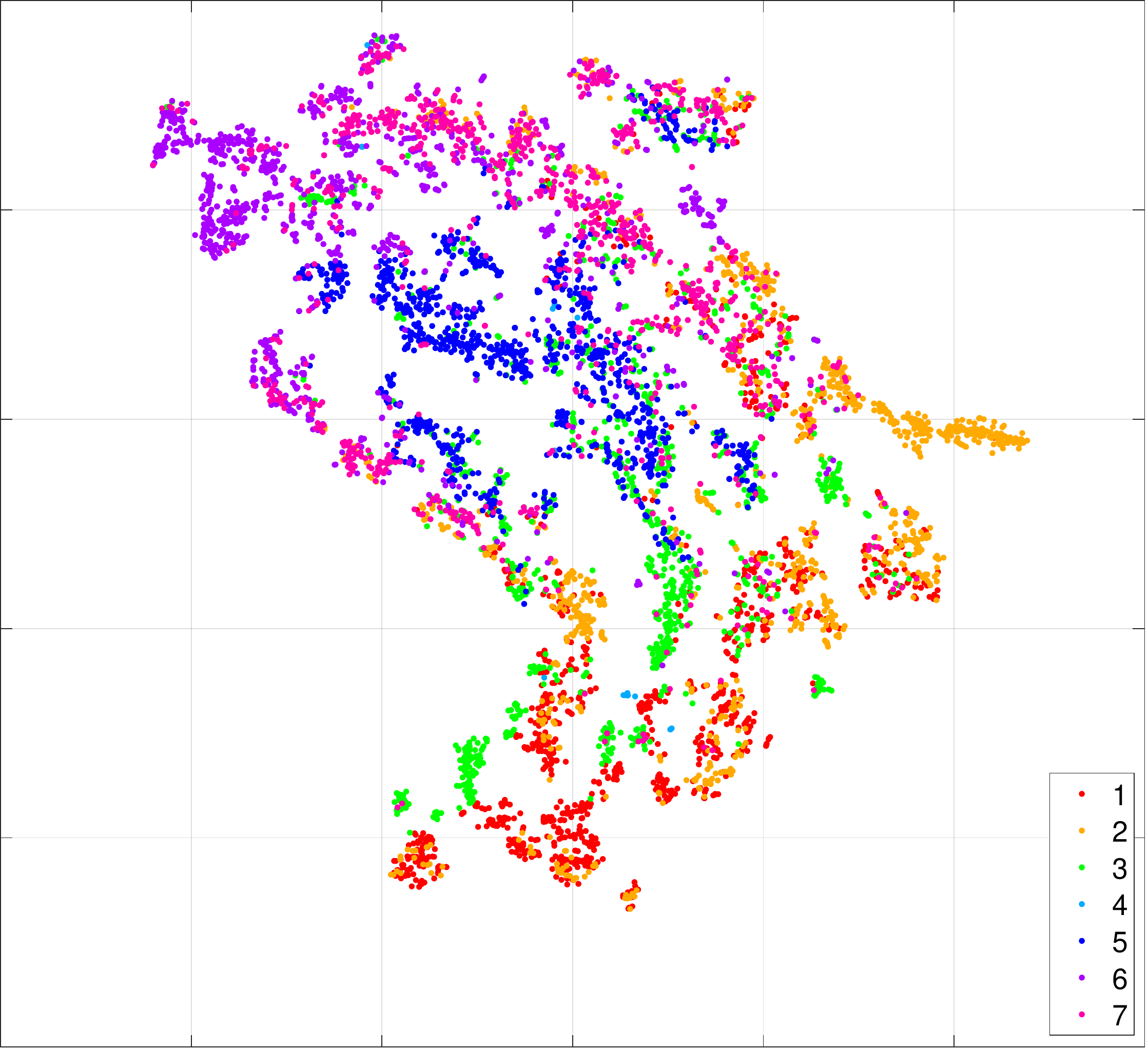} &
			\epsfig{width=1\figurewidth,file=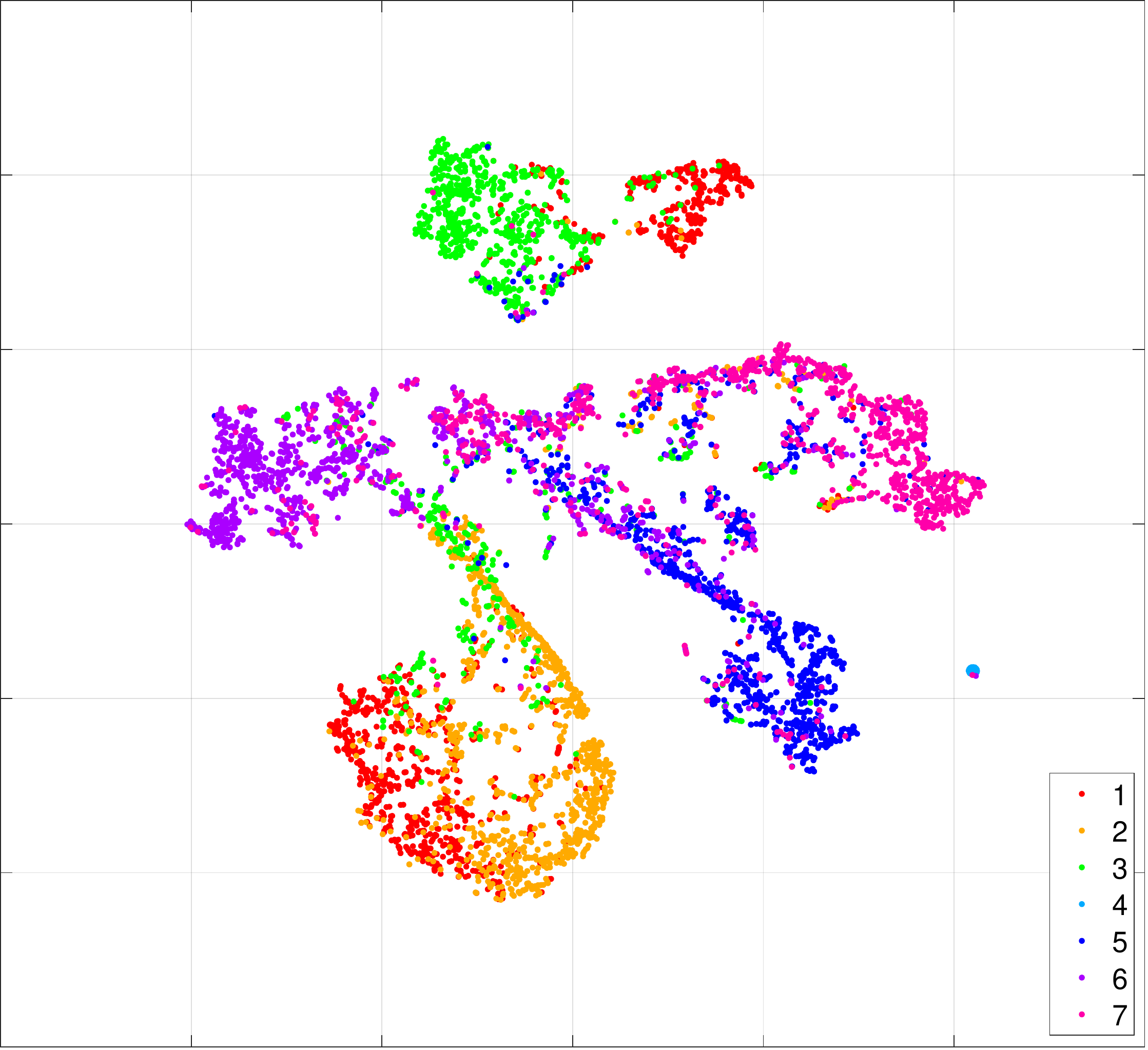} \\
			 (c) Original samples from TD & (d) TD Features by SDEnet 
			\\  [0.5em]
		\end{tabular}
		\caption{\label{fig:analysis}
			Class separability of the proposed SDEnet using the Houston dataset, where SD is the Houston 2013 data, TD is the Houston 2018 data, $\bullet$ represents SD, × represents ED, and the number represents class index. it is obvious that features by SDEnet have the best class separability.   }
	\end{center}
\vspace{-2em}
\end{figure*}

%

\subsection{Generalization Performance of SDEnet}
In SDEnet, the generator $G$ is required to produce ED with domain-specific information (domain shift with SD) and domain invariant discriminat information, while the discriminator $D$ is expected to extract class-wise domain invariant representations from SD, ID and ED. We take the Houston dataset as an example to illustrate $G$ and $D$. As shown in Fig. \ref{fig:analysis}(a) original samples from SD, and ED output by $G$, and (b) SD features and ED features output by $P$, where SD is the Houston 2013 data, $\bullet$ represents SD, × represents ED, and the number represents class index. It can be seen from Fig. \ref{fig:analysis}(a), for instance, in the 1-st class, 2-nd class and 3-rd class, there is a certain domain shift between SD and ED of the same class (the same color). Further, the samples of the same class in ED are aggregated better, indicating that they maintain the discriminant information in SD. It is obvious that $D$ extracts the class-wise domain invariant representation, as shown in Fig. \ref{fig:analysis}(b), where $D$ effectively alleviates the domain shift between SD and ED during training. In addition, the MMD distance between SD and ED of the original samples and features output by $P$ are quantitatively analyzed in Table \ref{table:analysis}. There is a distribution gap of about 0.78 between ED and SD in each class, and the gap is reduced by about 0.6 after obtaining embedding features through $D$. This also indicates that $G$ and $D$ in SDEnet perform their duties well and learn the class-wise domain invariant representation.

The TD (Houston 2018 data) is inferred directly by fully trained SDEnet. The visualization of class separability in the original space and feature space are shown in Fig. \ref{fig:analysis}(c) and (d). The inter-class distribution is mixed together in Fig. \ref{fig:analysis}(c), and the separability is significantly improved after feature embedding by SDEnet. In Fig. \ref{fig:analysis}(d), the inter-class distance is obviously increased and the intra-class distance is reduced. This shows that SDEnet has the ability to extract class-wise domain invariant representations and is well generalized to TD by learning the differences between SD, ID and ED. In addition, it can be seen from the $mmd(SD,TD)$ in Table \ref{table:analysis} that the original distribution gap between SD and TD is also decreased by about 0.1 in the feature space of SDEnet output.
\begin{table}[]
		\caption{\label{table:analysis}
	The MMD distance between SD and ED and between SD and TD of the original samples and features output by $P$. \\ (Houston dataset)}
	\begin{center}
	\begin{tabular}{|c|cc|cc|}
		\hline \hline
		\multirow{2}{*}{Class} & \multicolumn{2}{c|}{$mmd(SD,ED)$}              & \multicolumn{2}{c|}{$mmd(SD,TD)$}              \\ \cline{2-5} 
		& \multicolumn{1}{c|}{origin} & projection & \multicolumn{1}{c|}{origin} & projection \\ \hline
		1                    & \multicolumn{1}{c|}{0.7881} & 0.1671     & \multicolumn{1}{c|}{0.4870} & 0.3723     \\
		2                    & \multicolumn{1}{c|}{0.7854} & 0.1753     & \multicolumn{1}{c|}{0.4386} & 0.3189     \\
		3                    & \multicolumn{1}{c|}{0.7882} & 0.1620     & \multicolumn{1}{c|}{0.4126} & 0.3239     \\
		4                    & \multicolumn{1}{c|}{0.7579} & 0.1819     & \multicolumn{1}{c|}{0.3610} & 0.2909     \\
		5                    & \multicolumn{1}{c|}{0.8056} & 0.1729     & \multicolumn{1}{c|}{0.4657} & 0.3474     \\
		6                    & \multicolumn{1}{c|}{0.7661} & 0.1789     & \multicolumn{1}{c|}{0.4088} & 0.3193     \\
		7                    & \multicolumn{1}{c|}{0.7863} & 0.1758     & \multicolumn{1}{c|}{0.4212} & 0.3301     \\ \hline
		Mean                 & \multicolumn{1}{c|}{0.7825} & \textbf{0.1734}     & \multicolumn{1}{c|}{0.4278} & \textbf{0.3290}     \\ \hline \hline
	\end{tabular}
	\end{center}
\vspace{-2em}
\end{table}

\section{Experimental Results and Discussion}
\label{sec:results}
Experiments using three cross-scene HSI datasets, i.e., the Houston dataset, Pavia dataset, and GID (Gaofen Image Dataset) dataset, are conducted to validate the proposed SDEnet. Several state-of-the-art transfer learning algorithms are employed for comparison algorithms, including  DA methods, Dynamic Adversarial Adaptation Network (DAAN) \cite{yu2019transfer}, Deep Subdomain Adaption Network (DSAN) \cite{2020Deepsub}, Multi-Representation Adaptation Network (MRAN) \cite{ZHU2019214} and Heterogeneous Transfer CNN (HTCNN) \cite{2019Heterogeneous}, DG methods, Progressive Domain Expansion Network (PDEN) \cite{li2021progressive}, LDSDG (Learning to Diversify for Single Domain Generalization) \cite{wang2021learning} and Style-Agnostic Network (SagNet) \cite{nam2021reducing}. The class-specific accuracy (CA), the overall accuracy (OA) and the Kappa coefficient (KC) are employed to evaluate the classification performance.

\subsection{Experimental Data}
\begin{figure*}[tp]
	\begin{center}
		\begin{tabular}{cc}
			\epsfig{width=\figurewidth,file=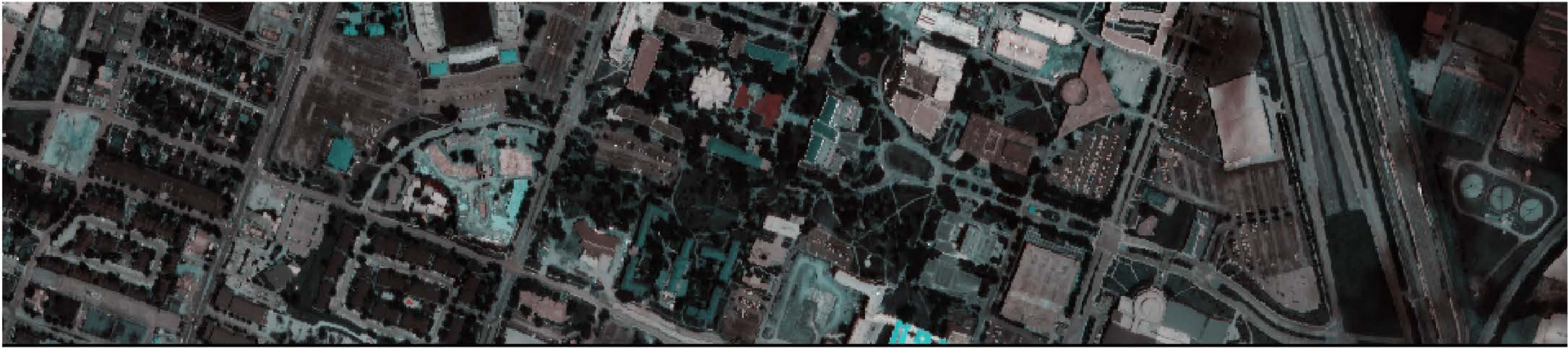} &
			\epsfig{width=\figurewidth,file=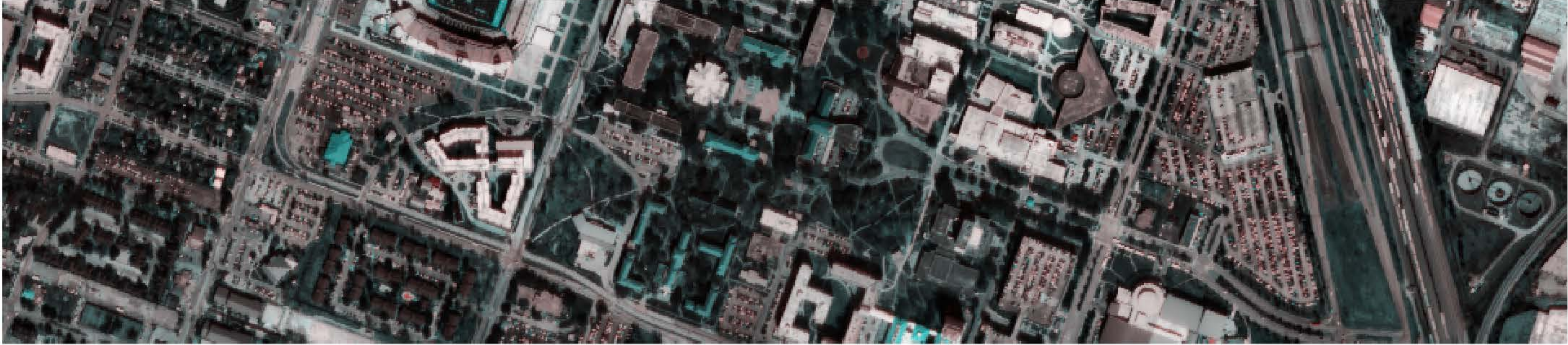} \\
			(a) & (b) \\
			\epsfig{width=\figurewidth,file=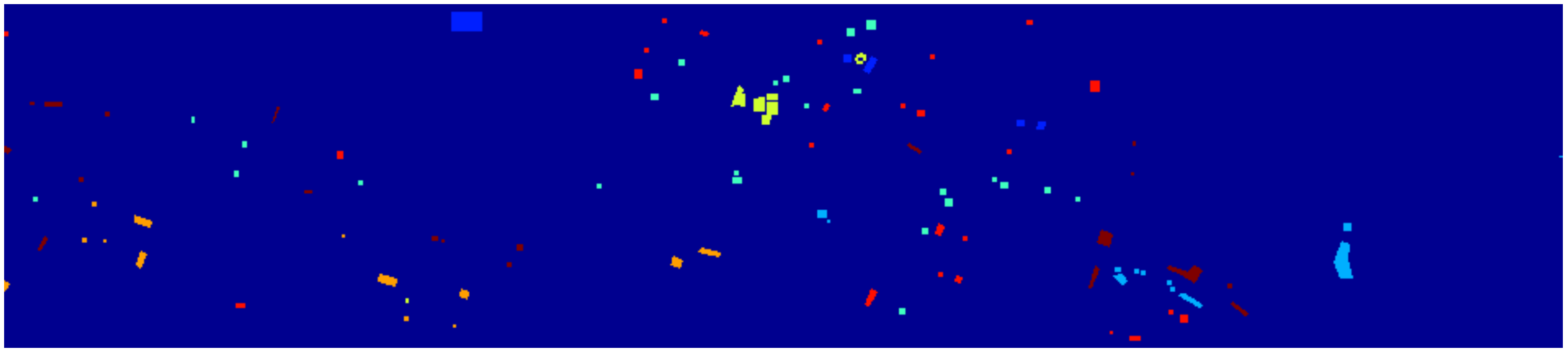} &
			\epsfig{width=\figurewidth,file=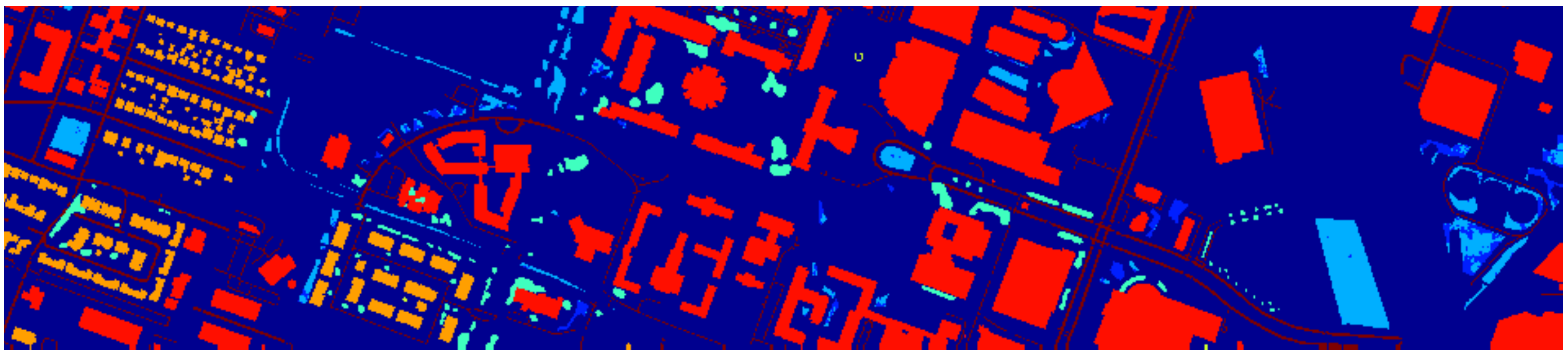} \\
			(c) & (d)   \\
		\end{tabular}
		\begin{tabular}{cc}
			\epsfig{width=\figurewidth,file=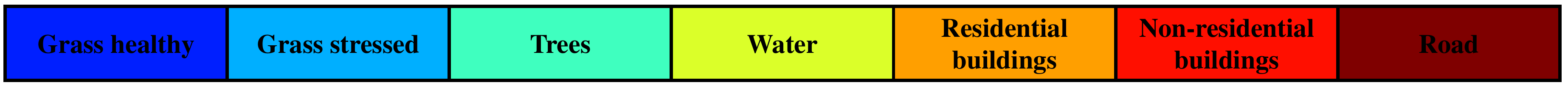}\\
		\end{tabular}
	\end{center}
	\caption{\label{fig:Houston_fg}
		Pseudo-color image and ground truth map of Houston dataset: (a) Pseudo-color image of Houston 2013, (b) Pseudo-color image of Houston 2018, (c) Ground truth map of Houston 2013, (d) Ground truth map of Houston 2018.}
\end{figure*}

\textbf{Houston dataset}: The dataset includes Houston 2013 \cite{2014Hyperspectral} and Houston 2018 \cite{20182018} scenes, which were obtained by different sensors on the University of Houston campus and its vicinity in different years. The Houston 2013 dataset is composed of 349$\times$1905 pixels, including 144 spectral bands, the wavelength range is 380-1050nm, and the image spatial resolution is 2.5m. The Houston 2018 dataset has the same wavelength range but contains 48 spectral bands, and the image has a spatial resolution of 1m. There are seven consistent classes in their scene. We extract 48 spectral bands (wavelength range 0.38$\sim$1.05um) from Houston 2013 scene corresponding to Houston 2018 scene, and select the overlapping area of 209$\times$955. The  classes and the number of samples are listed in Table \ref{table:Houston_samples}. Additionally, their false-color and ground truth maps are shown in Fig. \ref{fig:Houston_fg}.

\begin{figure*}[tp]
	\begin{center}
		\begin{tabular}{cccc}
			\epsfig{width=0.4\figurewidth,file=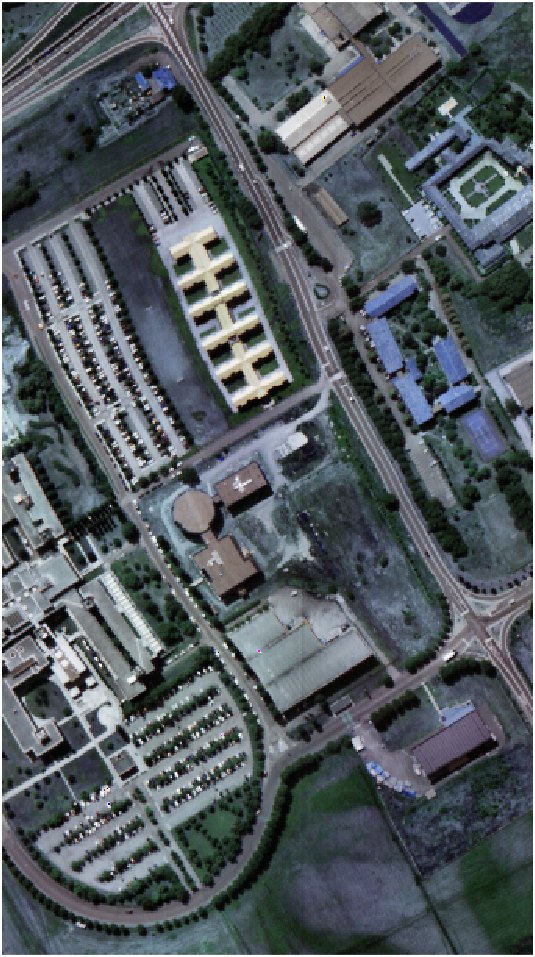}&
			\epsfig{width=0.405\figurewidth,file=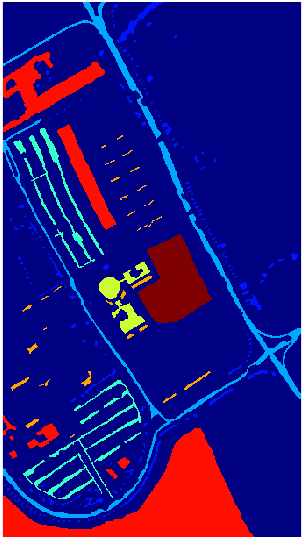}&
			\epsfig{width=0.45\figurewidth,file=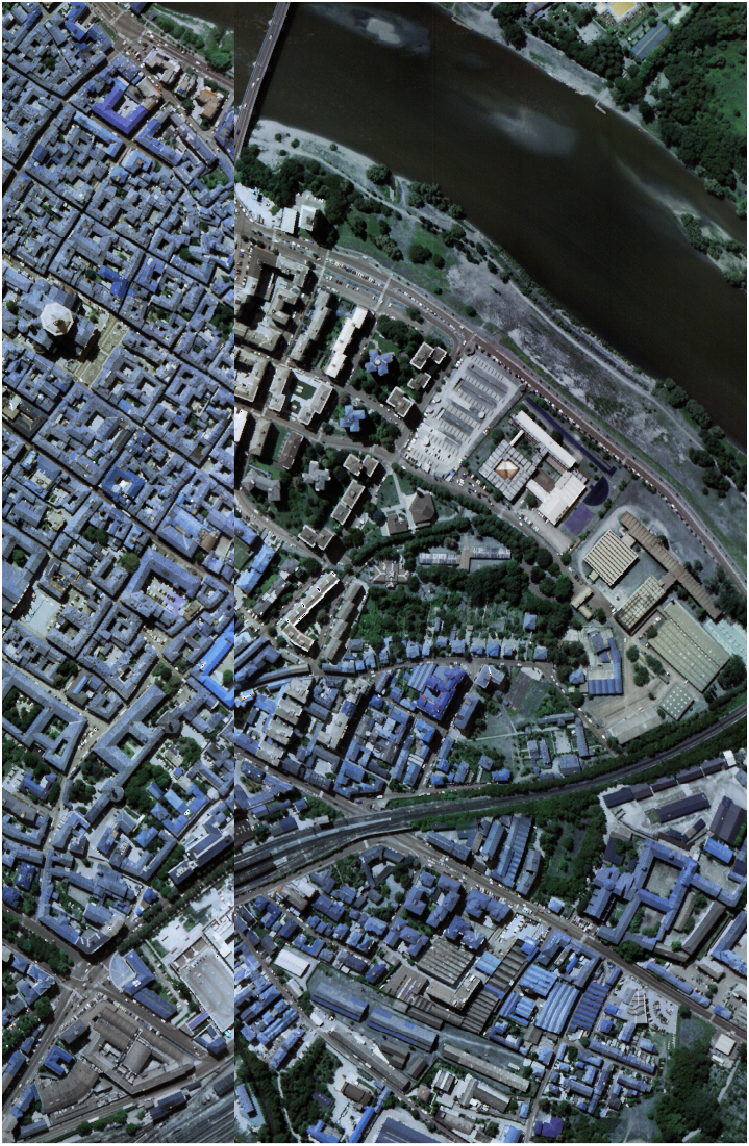} &
			\epsfig{width=0.45\figurewidth,file=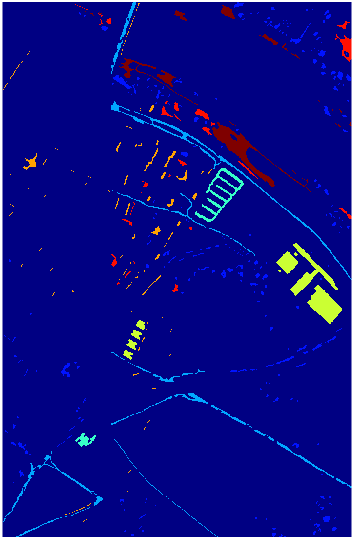} \\
			(a) & (b) & (c) & (d)  
		\end{tabular}
		\begin{tabular}{cc}
		\epsfig{width=1.5\figurewidth,file=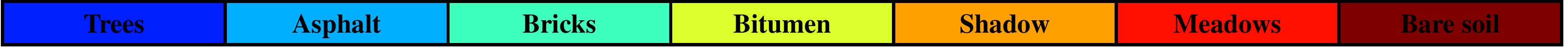}\\
		\end{tabular}
	\end{center}
	\caption{\label{fig:pavia_fg}
		Pseudo-color image and ground truth map of Pavia dataset: (a) Pseudo-color image of University of Pavia, (b) Ground truth map of University of Pavia, (c) Pseudo-color image of Pavia Center, (d) Ground truth map of Pavia Center.}
\end{figure*}

\textbf{Pavia dataset}: The Pavia dataset include University of Pavia (UP) and Pavia Center (PC). Both were gathered by Reflective Optics Spectrographic Image System (ROSIS), with spectral coverage 430 nm to 860nm. The PC has 1096$\times$715 pixels and 102 bands. The UP has 103 spectral bands, 610$\times$340 pixels and 1.3 m spatial resolution, where the last band was removed to ensure the same number of spectral bands as PC. They all have the same seven classes and the name of land cover classes and the number of samples are listed in Table \ref{table:pavia_samples}. Fig. \ref{fig:pavia_fg} shows their false-color images and ground-truth maps.

\begin{figure*}[tp]
	\begin{center}
		\begin{tabular}{cccc}
			\begin{sideways}\epsfig{width=0.7\figurewidth,file=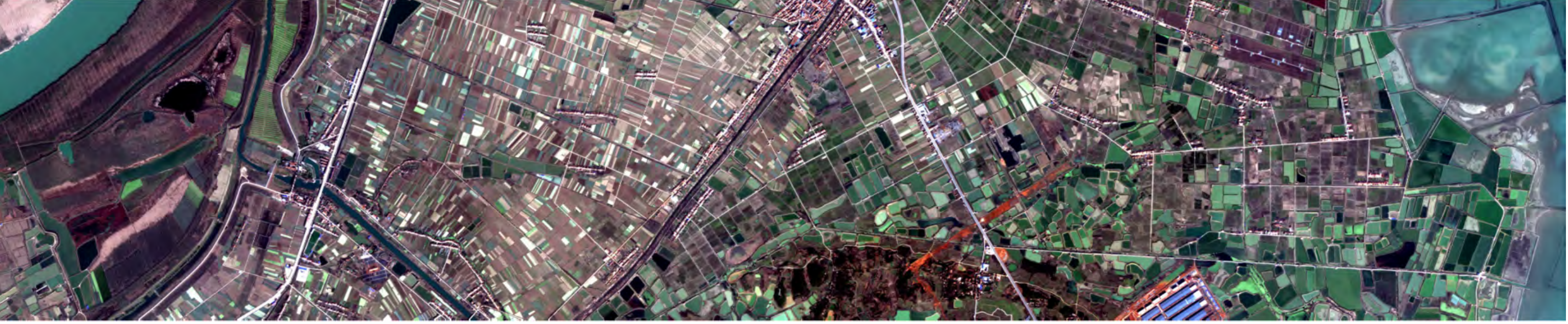}\end{sideways} &			\begin{sideways}\epsfig{width=0.7\figurewidth,file=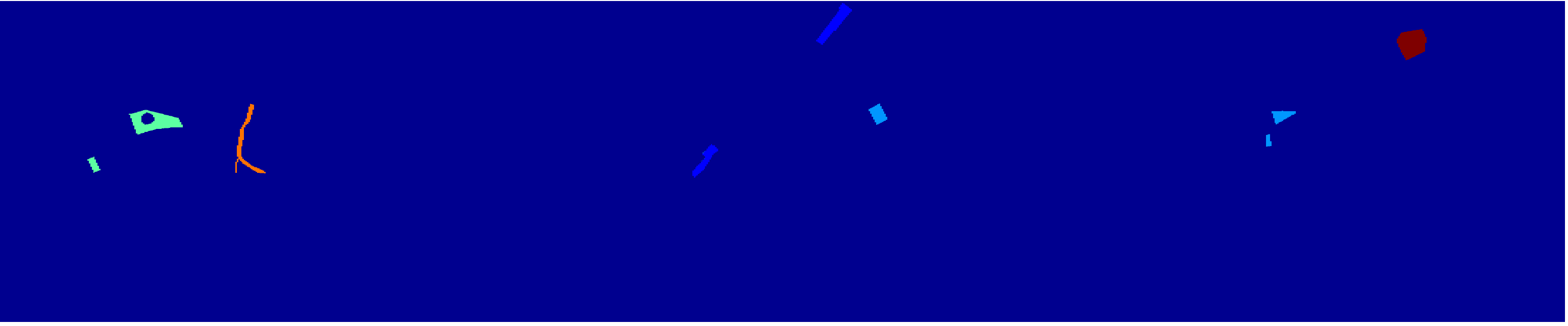} \end{sideways}&
			\epsfig{width=0.65\figurewidth,file=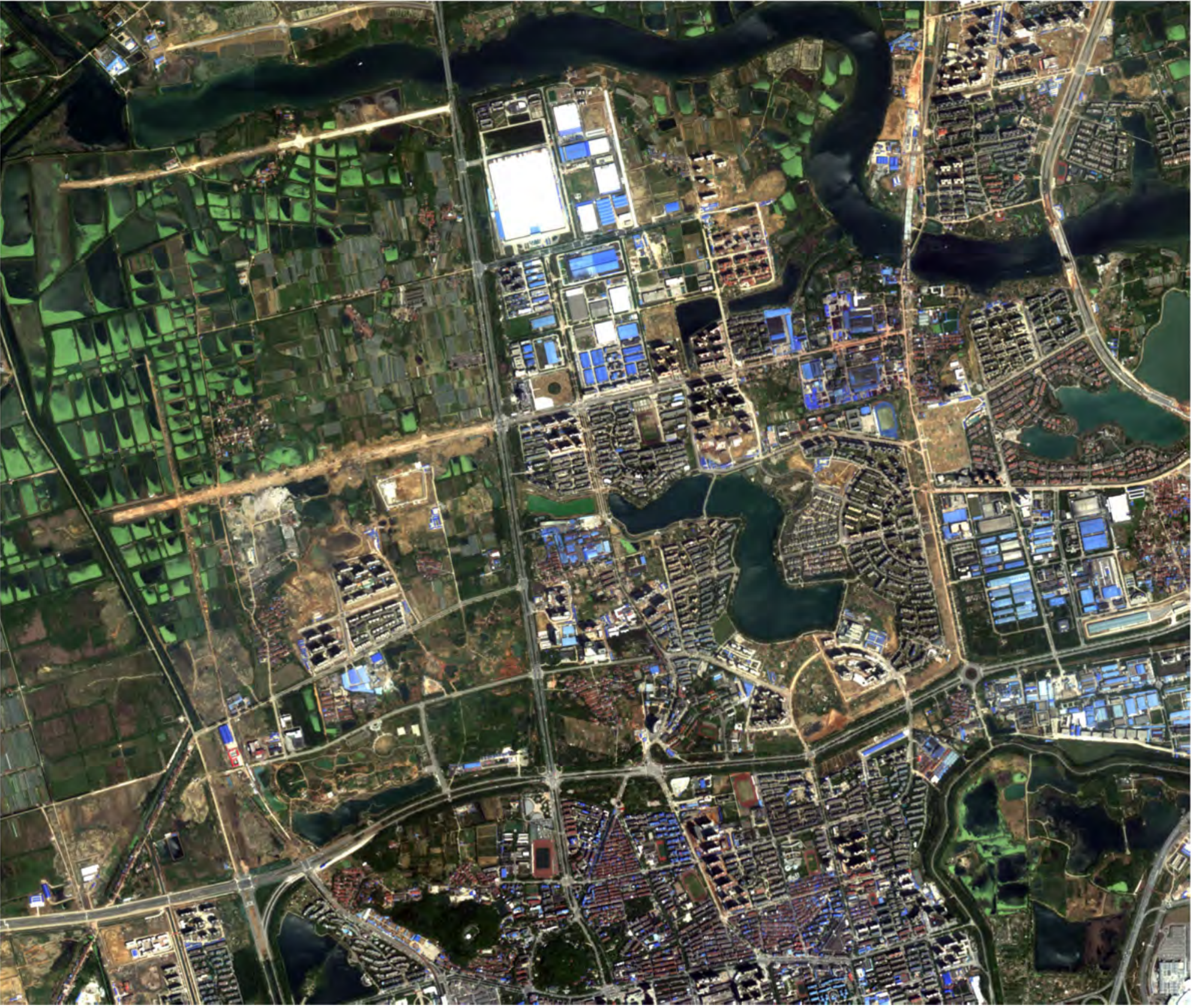} &
			\epsfig{width=0.65\figurewidth,file=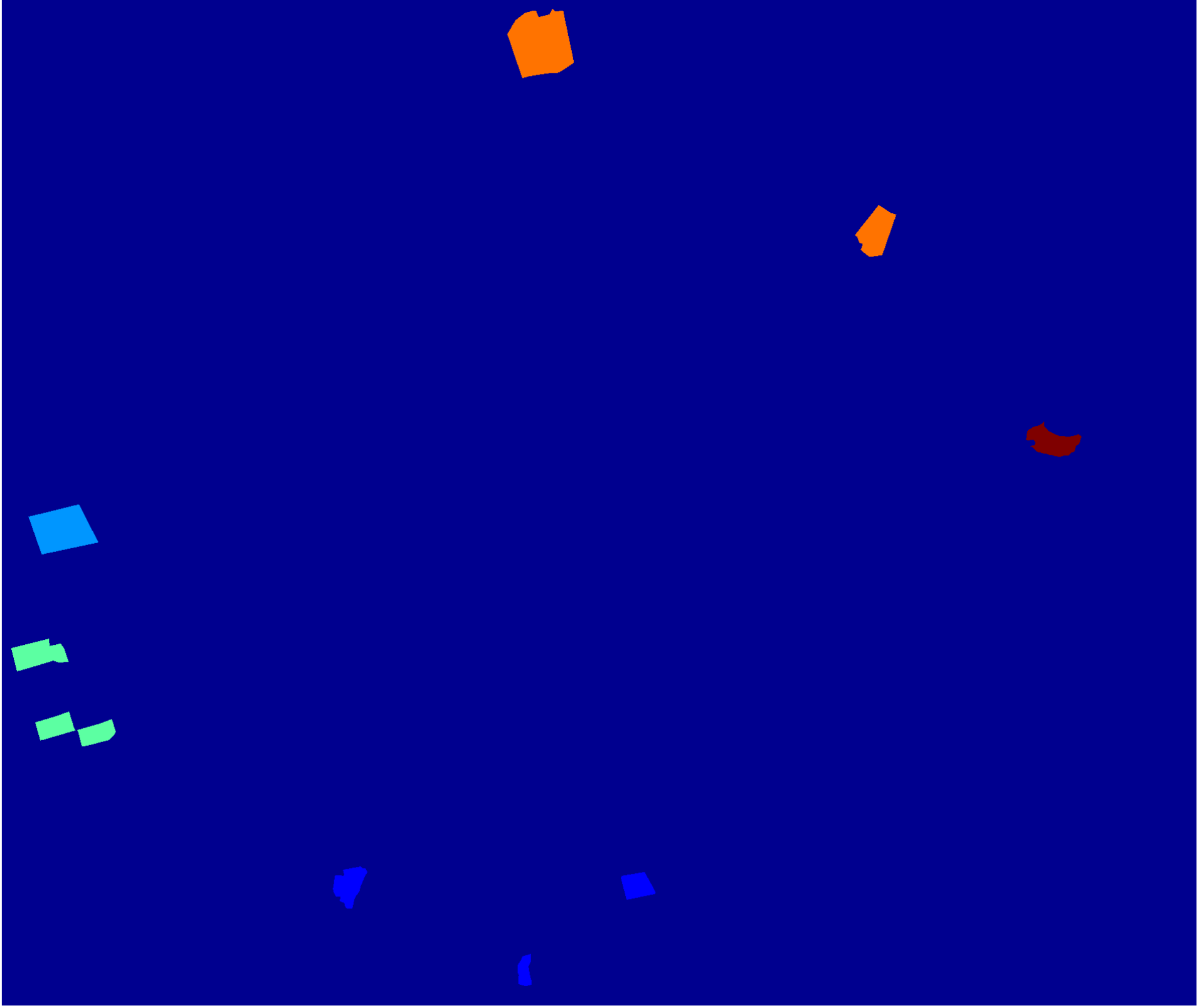} \\
				(a) & (b) & (c) & (d)  
		\end{tabular}
		\begin{tabular}{cc}
			\epsfig{width=1\figurewidth,file=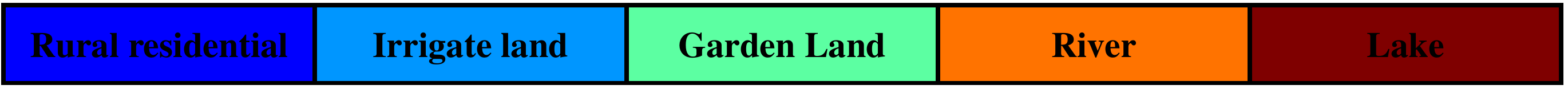}\\
		\end{tabular}
	\end{center}
	\caption{\label{fig:GID_fg}
		Pseudo-color image and ground truth map of GID dataset: (a) Pseudo-color image of GID-nc, (b) Ground truth map of GID-nc, (c) Pseudo-color image of GID-wh, (d) Ground truth map of GID-wh.}
\end{figure*}

\textbf{GID dataset}: GID dataset is constructed by Wuhan University \cite{tong2020land}, which contains multispectral images (MSI) taken at different times in many regions of China. The data comes from GF-2, which is the second satellite of the High Definition Earth Observation System (HDEOS) launched by China National Space Administration. We selected GID-nc shot in Nanchang, Jiangxi Province, on January 3, 2015 as the source domain, and GID-wh shot in Wuhan, Hubei Province, on April 11, 2016 as the target domain. GID-nc consists of 900$\times$4400 pixels, including blue (0.45-0.52um), green (0.52-0.59um), red (0.63-0.69um) and near infrared (0.77-0.89um) bands, and the spatial resolution is 4m. GID-wh also has the same spatial and spectral resolution, but it is composed of 1600$\times$1900 pixels. They have the same five classes, as listed in Table \ref{table:GID}. The false-color images and ground-truth maps are shown in Fig. \ref{fig:GID_fg}.

\begin{table}[tp]
	\caption{\label{table:Houston_samples}
		Number of source and target samples for the Houston dataset.}
	{
		\begin{center}
			\begin{tabular}{|c|c|c|c|}
				\hline
				\hline
				\multicolumn{2}{|c|}{Class} &\multicolumn{2}{c|}{Number of Samples} \\
				\hline
				\multirow{2}{*}{No.} &\multirow{2}{*}{Name}  & Houston 2013 & Houston 2018\\
				{~}  & {~}  & {(Source)} &  {(Target)} \\
				\hline
				1     & Grass healthy & 345   & 1353 \\
				2     & Grass stressed & 365   & 4888 \\
				3     & Trees & 365   & 2766 \\
				4     & Water & 285   & 22 \\
				5     & Residential buildings & 319   & 5347 \\
				6     & Non-residential buildings & 408   & 32459 \\
				7     & Road  & 443   & 6365 \\
				\hline
				\multicolumn{2}{|c|}{Total} & 2530& 53200\\
				\hline \hline
			\end{tabular}
	\end{center}}
\vspace{-2em}
\end{table}

\begin{table}[tp]
	\caption{\label{table:pavia_samples}
		Number of source and target samples for the Pavia dataset.}
	{
		\begin{center}
			\begin{tabular}{|c|c|c|c|}
				\hline
				\hline
				\multicolumn{2}{|c|}{Class} &\multicolumn{2}{c|}{Number of Samples} \\
				\hline
				\multirow{2}{*}{No.} &\multirow{2}{*}{Name}  & UP & PC\\
				{~}  & {~}  & {(Source)} &  {(Target)} \\
				\hline
				1 & Tree      & 3064  & 7598 \\
				2 & Asphalt   & 6631  & 9248 \\
				3 & Brick     & 3682  & 2685 \\
				4 & Bitumen   & 1330  & 7287 \\
				5 & Shadow    & 947   & 2863 \\
				6 & Meadow    & 18649 & 3090 \\
				7 & Bare soil & 5029  & 6584 \\
				\hline
				\multicolumn{2}{|c|}{Total} & 39332& 39355\\
				\hline \hline
			\end{tabular}
	\end{center}}
\end{table}

\begin{table}[tp]
	\caption{\label{table:GID}
		Number of source and target samples for the GID dataset.}
	{
		\begin{center}
			\begin{tabular}{|c|c|c|c|}
				\hline
				\hline
				\multicolumn{2}{|c|}{Class} &\multicolumn{2}{c|}{Number of Samples} \\
				\hline
				\multirow{2}{*}{No.} &\multirow{2}{*}{Name}  & GID-nc & GID-wh\\
				{~}  & {~}  & {(Source)} &  {(Target)} \\
				\hline
				1     & Rural residential & 5495  & 4729 \\
				2     & Irrigate land & 3643   & 5643 \\
				3     & Garden Land & 6171   & 6216 \\
				4     & River& 2858   & 11558 \\
				5     & Lake & 5172  & 2666 \\
				\hline
				\multicolumn{2}{|c|}{Total} & 23339 & 30812\\
				\hline \hline
			\end{tabular}
	\end{center}}
\end{table}

%
%
%
%


\begin{table}[]
	\caption{\label{tab:lr}
		Parameter tuning of the base learning rate ${\eta}$ for the proposed SDEnet using the three experimental data.}
	\centering
	\begin{tabular}{|c|c|c|c|c|c|}
		\hline\hline
		\multirow{2}{*}{Target scene} & \multicolumn{5}{c|}{Base learning rate   ${\eta}$}             \\ \cline{2-6}
		& 1e-5           & 1e-4  & 1e-3           & 1e-2           & 1e-1  \\ \hline
		Houston & 54.84 & 71.99 & \textbf{79.96} & 64.92 & 61.04 \\ \hline
		Pavia   & 62.06 & 81.82 & \textbf{81.94} & 57.56 & 40.21 \\ \hline
		GID     & 74.27 & 71.49 & \textbf{77.73} & 63.06 & 54.89 \\ \hline\hline
	\end{tabular}
\vspace{-0.1em}
\end{table}

\begin{table}[]
	\caption{\label{tab:lambda}
		Parameter tuning of the regularization parameter ${\lambda}$ for the proposed SDEnet using the three experimental data.}
	\centering
	\begin{tabular}{|c|c|c|c|c|c|}
		\hline\hline
		\multirow{2}{*}{Target scene} & \multicolumn{5}{c|}{Regularization parameter ${\lambda}$}             \\ \cline{2-6}
		& 1e-3           & 1e-2  & 1e-1           & 1e+0           & 1e+1  \\ \hline
		Houston      & 77.82          & 76.58 & \textbf{79.96} & 77.43 & 75.77 \\ \hline
		Pavia        & 80.54          & 81.27 & \textbf{81.94} & 80.57 & 80.14 \\ \hline
		GID & 69.59 & 69.00             & \textbf{77.73} & 73.46 & 76.38 \\ \hline\hline
	\end{tabular}
	\vspace{-0.1em}
\end{table}

\begin{table}[]
	\caption{\label{tab:dse}
		Parameter tuning of the embedding feature dimension ${d_{se}}$ for the proposed SDEnet using the three experimental data.}
	\centering
	\begin{tabular}{|c|c|c|c|c|}
		\hline\hline
		\multirow{2}{*}{Target scene} & \multicolumn{4}{c|}{Embedding feature dimension ${d_{se}}$ }             \\ \cline{2-5}
		& 16          & 32 & 64           & 128    \\ \hline
		Houston      & 77.17          & 78.73 & \textbf{79.96} & 78.61 \\ \hline
		Pavia        & 77.32          &77.47 & \textbf{81.94} & 80.92 \\ \hline
		GID & \textbf{77.73} & 75.74          & 63.17          & 62.89 \\ \hline\hline
	\end{tabular}
	\vspace{-0.1em}
\end{table}


\begin{table}[tp]
	\begin{center}
		\centering
		\setlength{\tabcolsep}{1mm}
		\caption{\label{table:Ablation}
			Ablation comparison of each variant of SDEnet.}
		\begin{tabular}{|c|ccccc|}
			\hline
			\hline
		\multicolumn{1}{|c|}{Model}    & \multicolumn{1}{c|}{\begin{tabular}[c]{@{}c@{}}SDEnet\\ (no se)\end{tabular}}            & \multicolumn{1}{c|}{\begin{tabular}[c]{@{}c@{}}SDEnet\\ (no me)\end{tabular}} & \multicolumn{1}{c|}{\begin{tabular}[c]{@{}c@{}}SDEnet\\ (no con)\end{tabular}} & \multicolumn{1}{c|}{\begin{tabular}[c]{@{}c@{}}SDEnet\\ (no adv)\end{tabular}} & \multicolumn{1}{c|}{SDEnet} \\ \hline
		\multicolumn{1}{|c|}{Data set} & \multicolumn{5}{c|}{Houston}
		\\ \hline
		OA (\%)                        & \multicolumn{1}{c|}{76.84} & \multicolumn{1}{c|}{78.84}                                              & \multicolumn{1}{c|}{75.57}                                              & \multicolumn{1}{c|}{76.78}                                              & \textbf{79.96}                \\ \hline
		KC ($\kappa$)                  & \multicolumn{1}{c|}{59.88} & \multicolumn{1}{c|}{64.52}                                              & \multicolumn{1}{c|}{57.57}                                              & \multicolumn{1}{c|}{59.78}                                              & \textbf{65.15}                \\ \hline
		& \multicolumn{5}{c|}{Pavia}                                                                                                                                                                                                                                                                                            \\ \hline
		OA (\%)                        & \multicolumn{1}{c|}{81.35} & \multicolumn{1}{c|}{78.69}                                              & \multicolumn{1}{c|}{80.02}                                              & \multicolumn{1}{c|}{81.31}                                              & \textbf{81.94}                \\ \hline
		KC ($\kappa$)                  & \multicolumn{1}{c|}{77.67} & \multicolumn{1}{c|}{74.52}                                              & \multicolumn{1}{c|}{76.17}                                              & \multicolumn{1}{c|}{77.58}                                              &\textbf{78.33}                   \\ \hline
		& \multicolumn{5}{c|}{GID}                                                                                                                                                                                                                                                                                                 \\ \hline
            
		OA (\%)                        & \multicolumn{1}{c|}{70.07} & \multicolumn{1}{c|}{64.80}                                            & \multicolumn{1}{c|}{68.34}                               & \multicolumn{1}{c|}{73.29}                                              & \textbf{77.73}                \\ \hline
		KC ($\kappa$)                  & \multicolumn{1}{c|}{60.73} & \multicolumn{1}{c|}{54.24}                                              & \multicolumn{1}{c|}{58.31}                                              & \multicolumn{1}{c|}{64.78}                                              & \textbf{70.47}  \\
				\hline
		\hline
	\end{tabular}
\end{center}
\vspace{-0.1em}
\end{table}

\begin{table*}[tp]
	\caption{\label{tab:accuracy_Hou}
		Class-specific and overall classification accuracy (\%) of different methods for the target scene Houston 2018 data.}
	{ 
		\begin{center}
	\begin{tabular}{|c||c|c|c|c|c|c|c|c|}
		\hline
		\hline
		\multirow{2}{*}{Class}  &  \multicolumn{8}{|c|}{Classification algorithms} \\
		\cline{2-9}
		{~}  & {DAAN \cite{yu2019transfer}}&  {MRAN \cite{ZHU2019214}} & {DSAN \cite{2020Deepsub}} & {HTCNN \cite{2019Heterogeneous}}  & {PDEN \cite{li2021progressive}}  & {LDSDG \cite{wang2021learning}}  & {SagNet \cite{nam2021reducing}}  & {SDEnet} \\
				\hline	
			1 & 68.29 & 41.02 & 62.31 & 11.83 & 46.49 & 10.13 & 25.79 & 24.69          \\
			2                       & 77.80  & 76.94 & 77.50  & 70.11 & 77.60  & 62.97 & 62.79 & 84.98          \\
			3                       & 67.50  & 65.91 & 74.55 & 54.99 & 59.73 & 60.81 & 48.66 & 59.65          \\
			4                       & 100   & 100   & 100   & 54.55 & 100   & 81.82 & 81.82 & 100            \\
			5                       & 47.69 & 36.90  & 73.39 & 55.60  & 49.62 & 45.65 & 59.57 & 62.33          \\
			6                       & 79.49 & 82.68 & 86.84 & 92.85 & 84.98 & 89.22 & 89.28 & 90.54          \\
			7                       & 45.12 & 56.43 & 46.33 & 46.47 & 64.21 & 44.15 & 34.99 & 57.45          \\	
				\hline
				\multicolumn{1}{|c|}{OA (\%)}& 71.13 & 72.48 & 78.52 & 77.42 & 75.98 & 73.55 & 73.64 & \textbf{79.96$\pm$1.18}  \\
				\hline
				\multicolumn{1}{|c|}{KC ($\kappa$)}  & 54.93 & 55.83 & 64.45 & 59.94 & 56.12 & 55.17 & 55.32 & \textbf{65.15$\pm$2.24}\\
				\hline \hline				
			\end{tabular}
	\end{center}}
\vspace{-0.1em}
\end{table*}

\begin{table*}[tp]
	\caption{\label{tab:accuracy_UP}
		Class-specific and overall classification accuracy (\%) of different methods for the target scene Pavia Center data.}
	{ 
		\begin{center}
			\begin{tabular}{|c||c|c|c|c|c|c|c|c|}
				\hline
				\hline
				\multirow{2}{*}{Class}  &  \multicolumn{8}{|c|}{Classification algorithms} \\
				\cline{2-9}
				{~}  & {DAAN \cite{yu2019transfer}}&  {MRAN \cite{ZHU2019214}} & {DSAN \cite{2020Deepsub}} & {HTCNN \cite{2019Heterogeneous}}  & {PDEN \cite{li2021progressive}}  & {LDSDG \cite{wang2021learning}}  & {SagNet \cite{nam2021reducing}}  & {SDEnet} \\
				\hline	
				1 & 71.98 & 59.16 & 93.93 & 96.06 & 85.93 & 91.09 & 98.35 & 89.93          \\
				2                       & 78.98 & 85.15 & 79.8  & 57.70  & 88.56 & 73.51 & 59.76 & 81.22          \\
				3                       & 19.37 & 46.18 & 53.97 & 2.76  & 61.34 & 2.23  & 5.40   & 72.77          \\
				4                       & 58.67 & 69.58 & 75.75 & 93.25 & 85.49 & 71.72 & 87.03 & 82.54          \\
				5                       & 70.87 & 64.58 & 99.44 & 89.94 & 87.95 & 71.04 & 93.19 & 84.81          \\
				6                       & 83.07 & 89.22 & 74.43 & 70.97 & 79.26 & 57.12 & 49.81 & 75.11          \\
				7                       & 55.59 & 60.10  & 67.31 & 42.28 & 64.75 & 78.13 & 57.94 & 78.74          \\	
				\hline
				\multicolumn{1}{|c|}{OA (\%)}& 65.62 & 69.22 & 78.94 & 68.75 & 80.87 & 71.02 & 69.90  & \textbf{81.94$\pm$1.55}  \\
				\hline
				\multicolumn{1}{|c|}{KC ($\kappa$)}   & 58.85 & 63.35 & 74.90  & 62.60  & 77.02 & 64.62 & 63.44 & \textbf{78.33$\pm$2.47}\\
				\hline \hline				
			\end{tabular}
	\end{center}}
	\vspace{-0.1em}
\end{table*}

\begin{table*}[tp]
	\caption{\label{tab:accuracy_GID}
		Class-specific and overall classification accuracy (\%) of different methods for the target scene GID-wh data.}
	{ 
		\begin{center}
			\begin{tabular}{|c||c|c|c|c|c|c|c|c|}
				\hline
				\hline
				\multirow{2}{*}{Class}  &  \multicolumn{8}{|c|}{Classification algorithms} \\
				\cline{2-9}
				{~}  & {DAAN \cite{yu2019transfer}}&  {MRAN \cite{ZHU2019214}} & {DSAN \cite{2020Deepsub}} & {HTCNN \cite{2019Heterogeneous}}  & {PDEN \cite{li2021progressive}}  & {LDSDG \cite{wang2021learning}}  & {SagNet \cite{nam2021reducing}}  & {SDEnet} \\
				\hline	
				1 & 93.93 & 36.79 & 94.99 & 27.00    & 79.47 & 22.71 & 37.64 & 88.64          \\
				2                       & 87.67 & 78.93 & 91.33 & 100   & 93.64 & 76.93 & 98.60  & 18.45          \\
				3                       & 11.13 & 74.39 & 11.89 & 0.00     & 2.48  & 99.29 & 1.87  & 90.09          \\
				4                       & 77.85 & 69.06 & 90.21 & 92.63 & 81.91 & 88.13 & 89.40  & 93.96          \\
				5                       & 71.01 & 74.83 & 71.08 & 0.00     & 82.52 & 67.25 & 45.01 & 84.70           \\	
				\hline
				\multicolumn{1}{|c|}{OA (\%)}& 68.06 & 67.49 & 73.69 & 57.20  & 67.71 & 76.48 & 61.64 & \textbf{77.73$\pm$2.03} \\
				\hline
				\multicolumn{1}{|c|}{KC ($\kappa$)}   & 58.61 & 57.82 & 65.29 & 43.06 & 57.97 & 68.80  & 48.74 & \textbf{70.47$\pm$2.84}\\
				\hline \hline				
			\end{tabular}
	\end{center}}
	\vspace{-0.1em}
\end{table*}


\begin{figure}[htp]
	\centering
	\setlength{\tabcolsep}{0.5em}
	\begin{tabular}{ccccccccccccc}
		\multicolumn{3}{c}{\begin{sideways}\epsfig{width=0.11\figurewidth,file=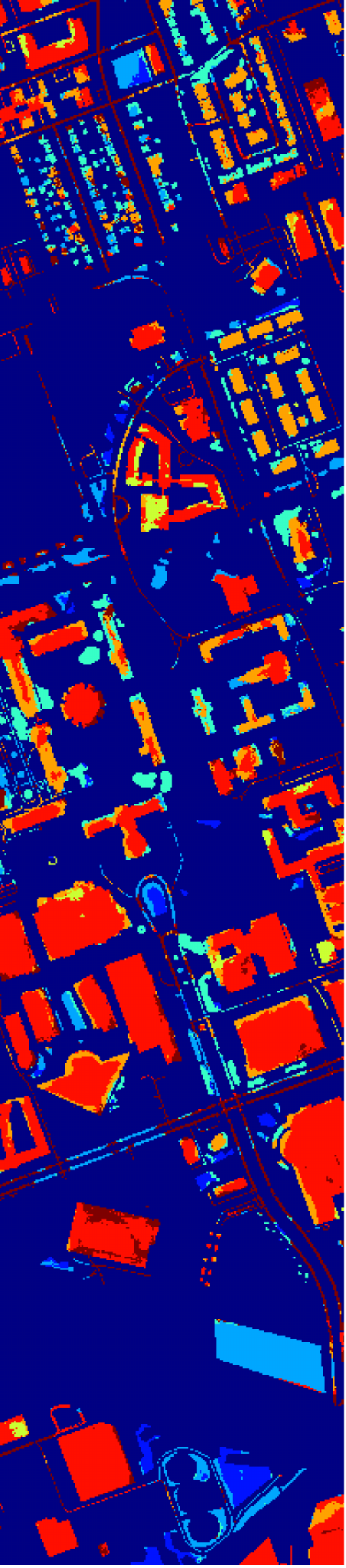}\end{sideways}}     &
		\multicolumn{3}{c}{\begin{sideways}\epsfig{width=0.11\figurewidth,file=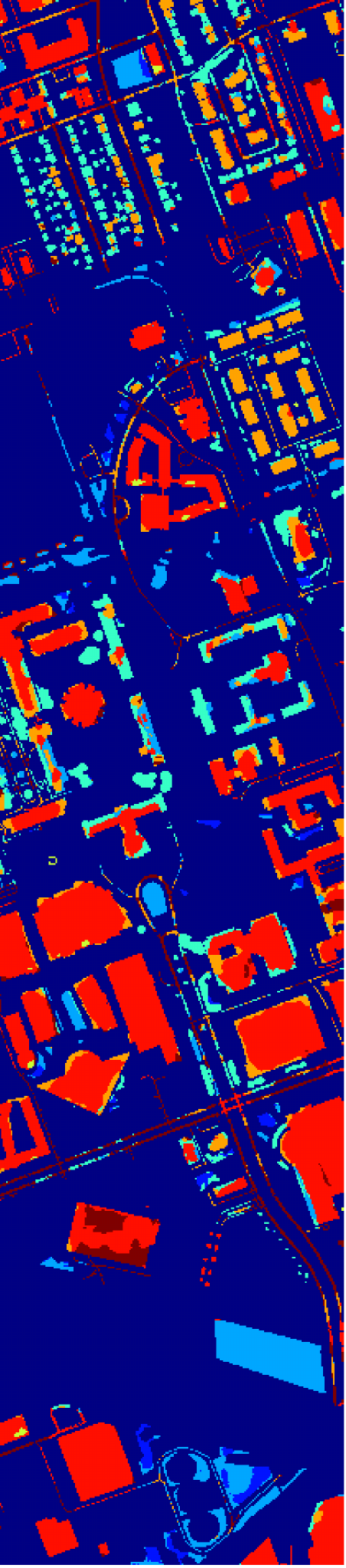}\end{sideways}}  \\
		\multicolumn{3}{c}{(a)} & \multicolumn{3}{c}{(b)} & \\
 		\multicolumn{3}{c}{\begin{sideways}\epsfig{width=0.11\figurewidth,file=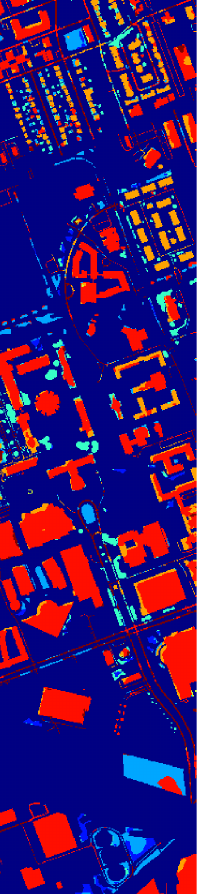}\end{sideways}}    &		\multicolumn{3}{c}{\begin{sideways}\epsfig{width=0.11\figurewidth,file=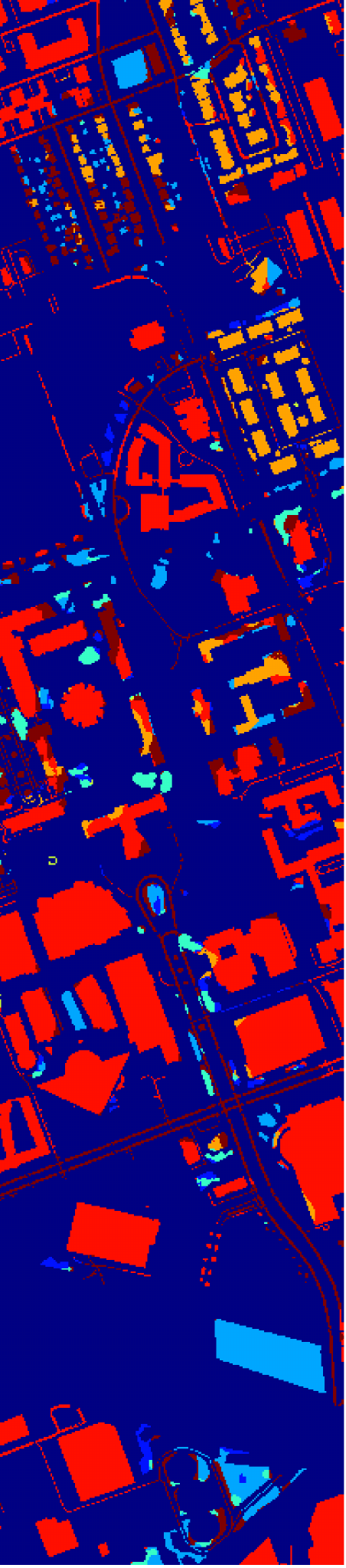}\end{sideways}}     \\
 		\multicolumn{3}{c}{(c)}  & \multicolumn{3}{c}{(d)} &    \\
		\multicolumn{3}{c}{\begin{sideways}\epsfig{width=0.11\figurewidth,file=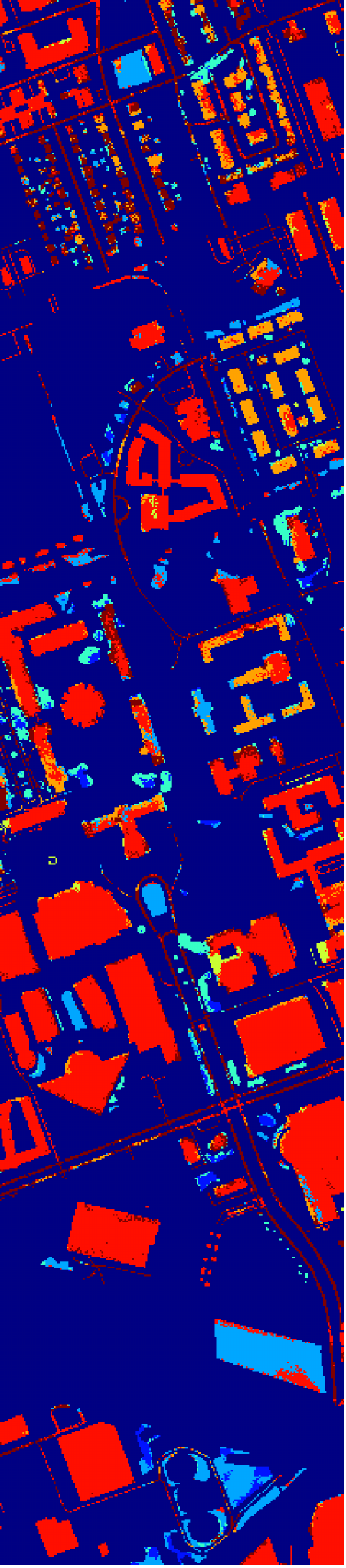}\end{sideways}}     &
		\multicolumn{3}{c}{\begin{sideways}\epsfig{width=0.11\figurewidth,file=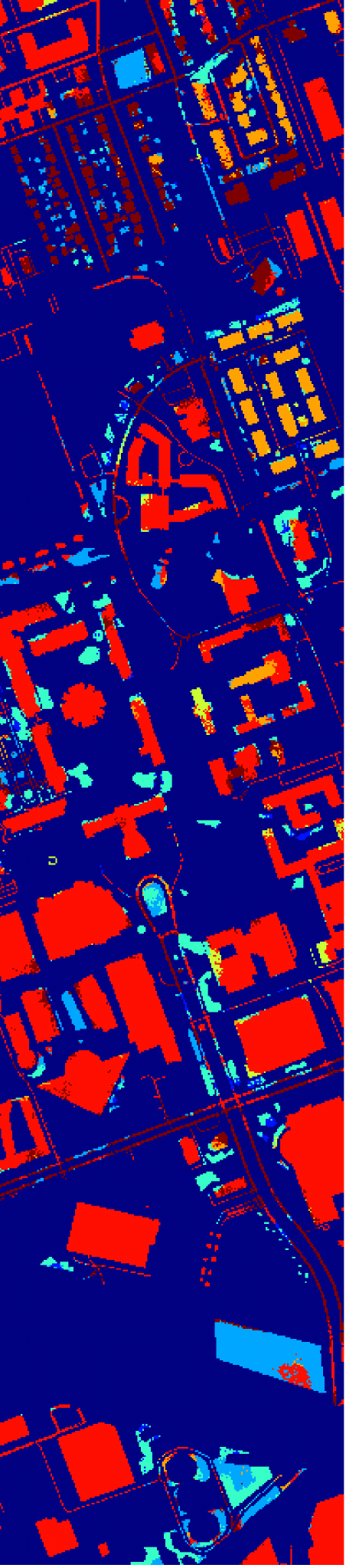}\end{sideways}}  \\
		\multicolumn{3}{c}{(e)} & \multicolumn{3}{c}{(f)} & \\
		\multicolumn{3}{c}{\begin{sideways}\epsfig{width=0.11\figurewidth,file=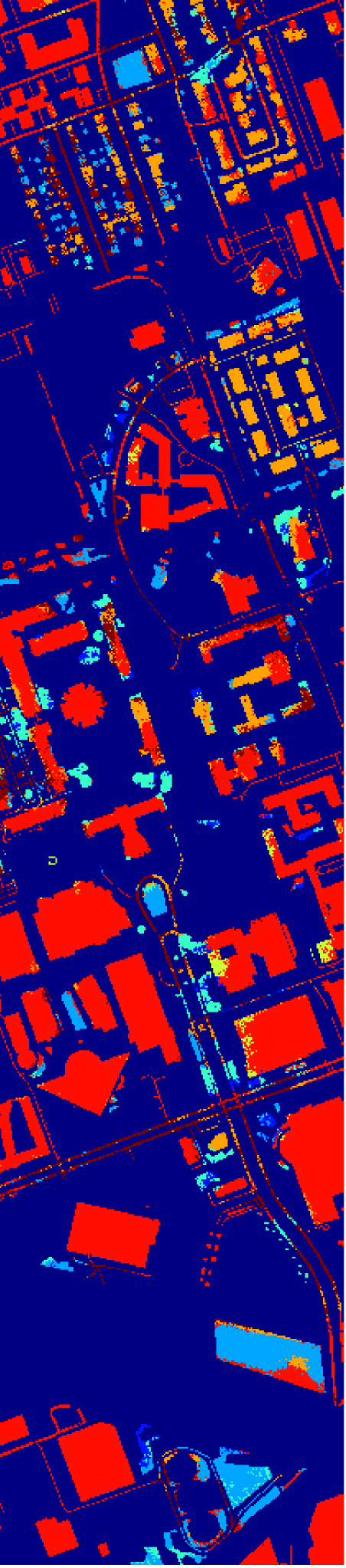}\end{sideways}}    &		\multicolumn{3}{c}{\begin{sideways}\epsfig{width=0.11\figurewidth,file=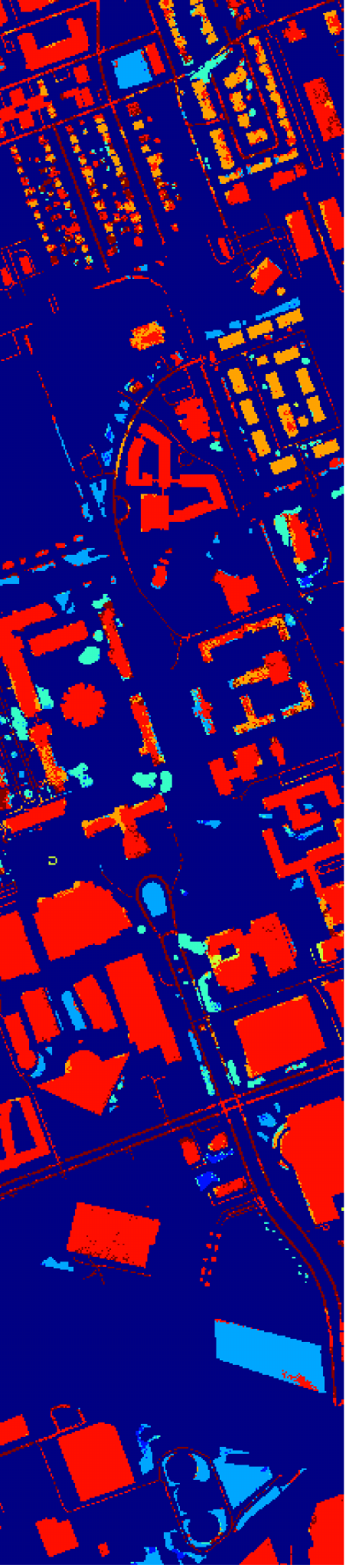}\end{sideways}}     \\
		\multicolumn{3}{c}{(g)}  & \multicolumn{3}{c}{(h)} &    \\
	\end{tabular}
	\vspace{-0.1em}
	\caption{\label{fig:Map_Hou}
		Data visualization and classification maps for target scene Houston 2018 data obtained with different methods including: (a) DAAN (71.13\%), (b) MRAN (72.48\%), (c) DSAN (78.52\%), (d) HTCNN (77.42\%), (e) PDEN (75.98\%), (f) LDSDG (73.55\%), (g) SagNet (73.64\%), (h)SDEnet (80.11\%).  }
\end{figure}

\begin{figure}[tp]
	\centering
	\vspace{-0.1em}
	\setlength{\tabcolsep}{0.1em}
	\begin{tabular}{ccccccccccccc}
		\multicolumn{3}{c}{\epsfig{width=0.25\figurewidth,file=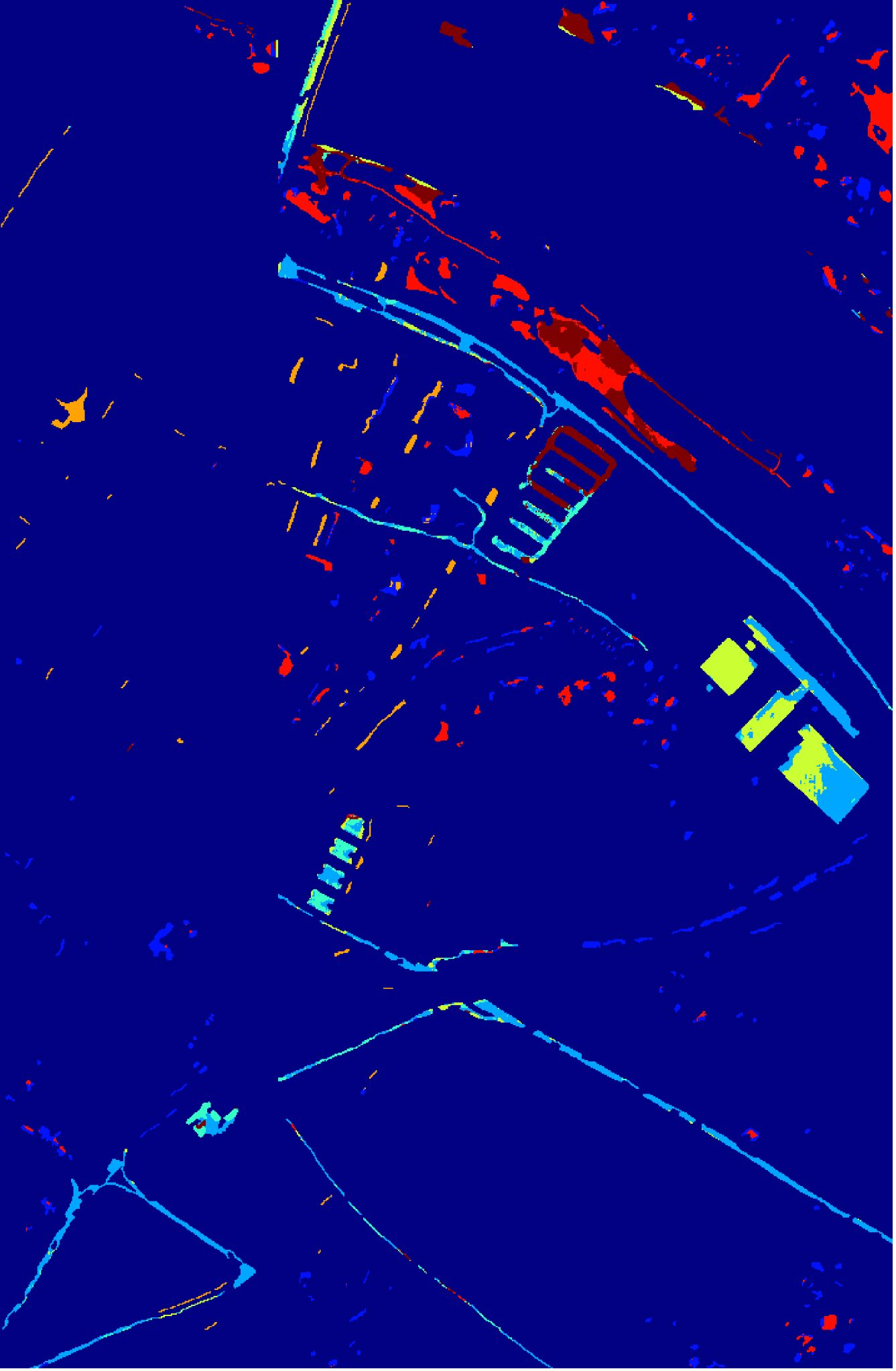}}     &
		\multicolumn{3}{c}{\epsfig{width=0.25\figurewidth,file=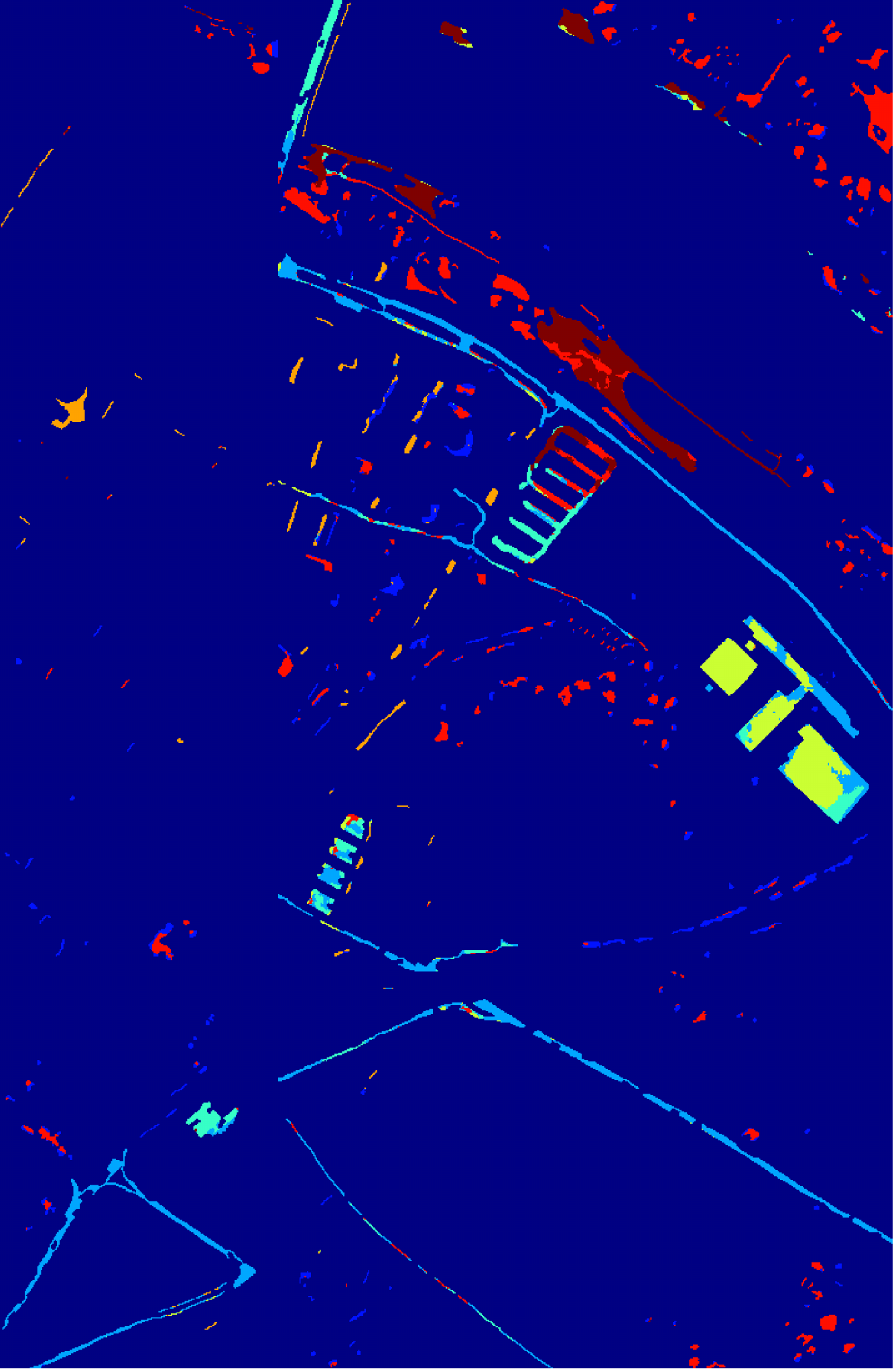}}  & 		\multicolumn{3}{c}{\epsfig{width=0.25\figurewidth,file=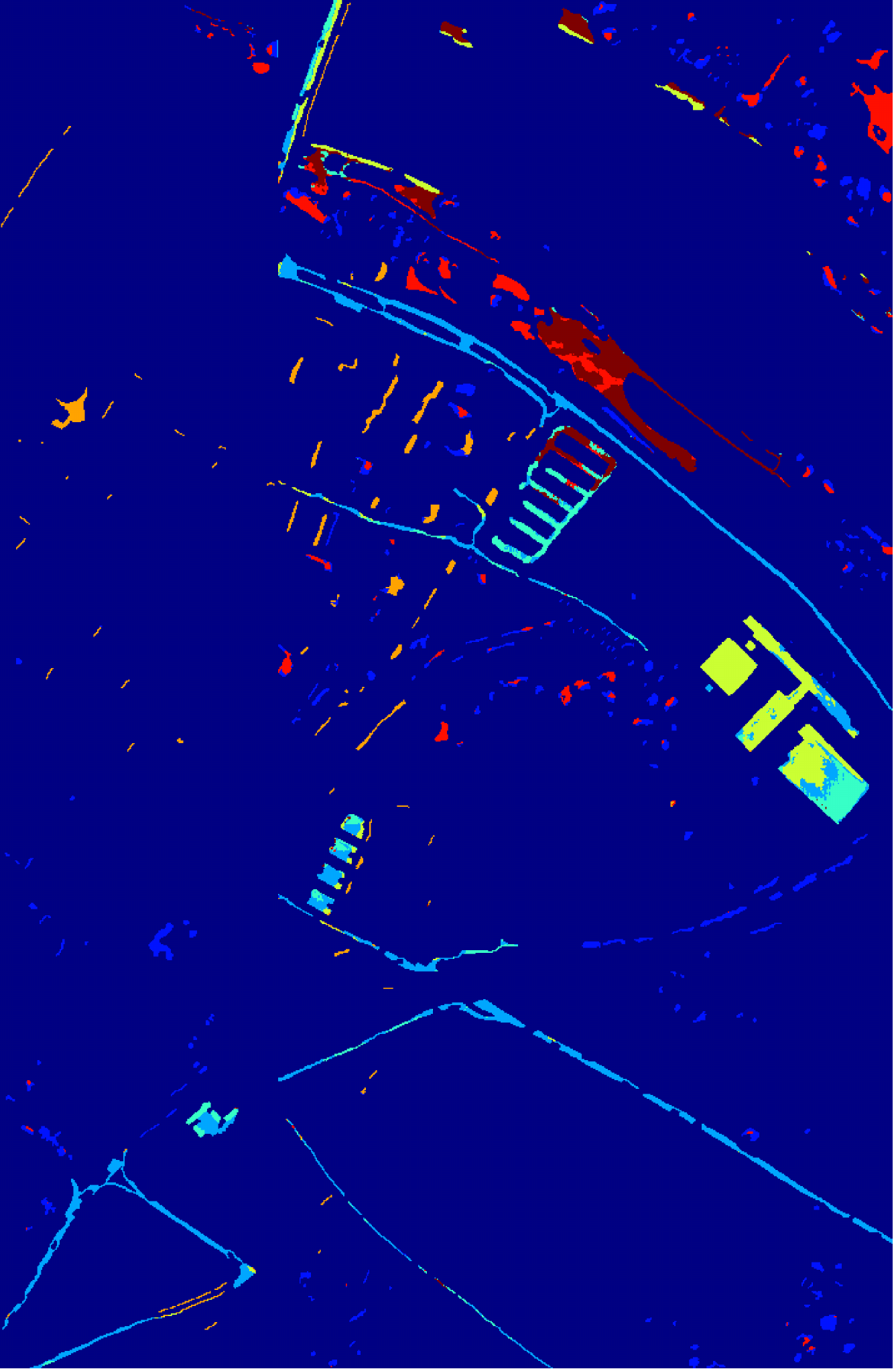}}    &		\multicolumn{3}{c}{\epsfig{width=0.25\figurewidth,file=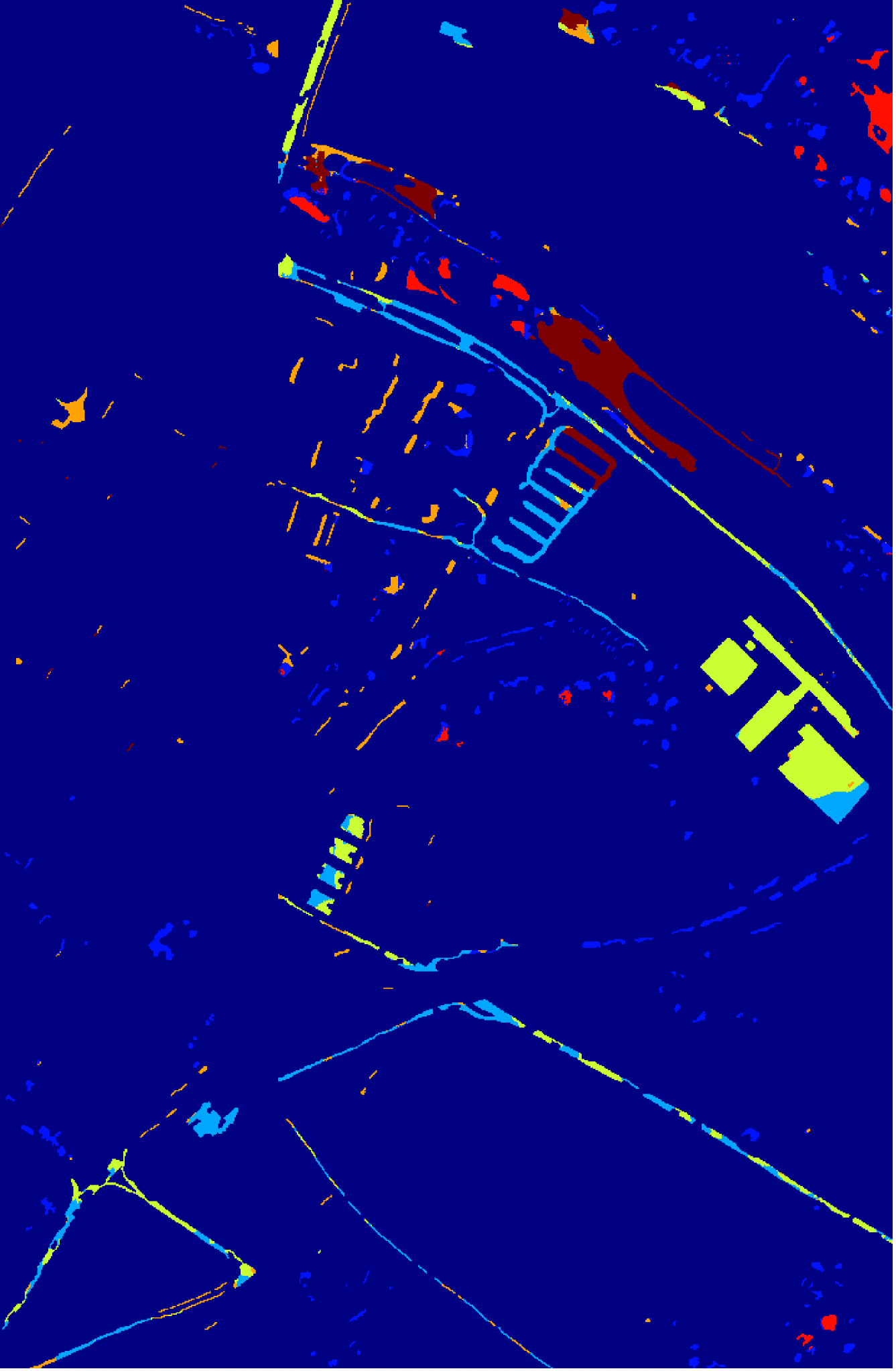}}  \\
		\multicolumn{3}{c}{(a)} & \multicolumn{3}{c}{(b)} & \multicolumn{3}{c}{(c)}  & \multicolumn{3}{c}{(d)} \\
		\multicolumn{3}{c}{\epsfig{width=0.25\figurewidth,file=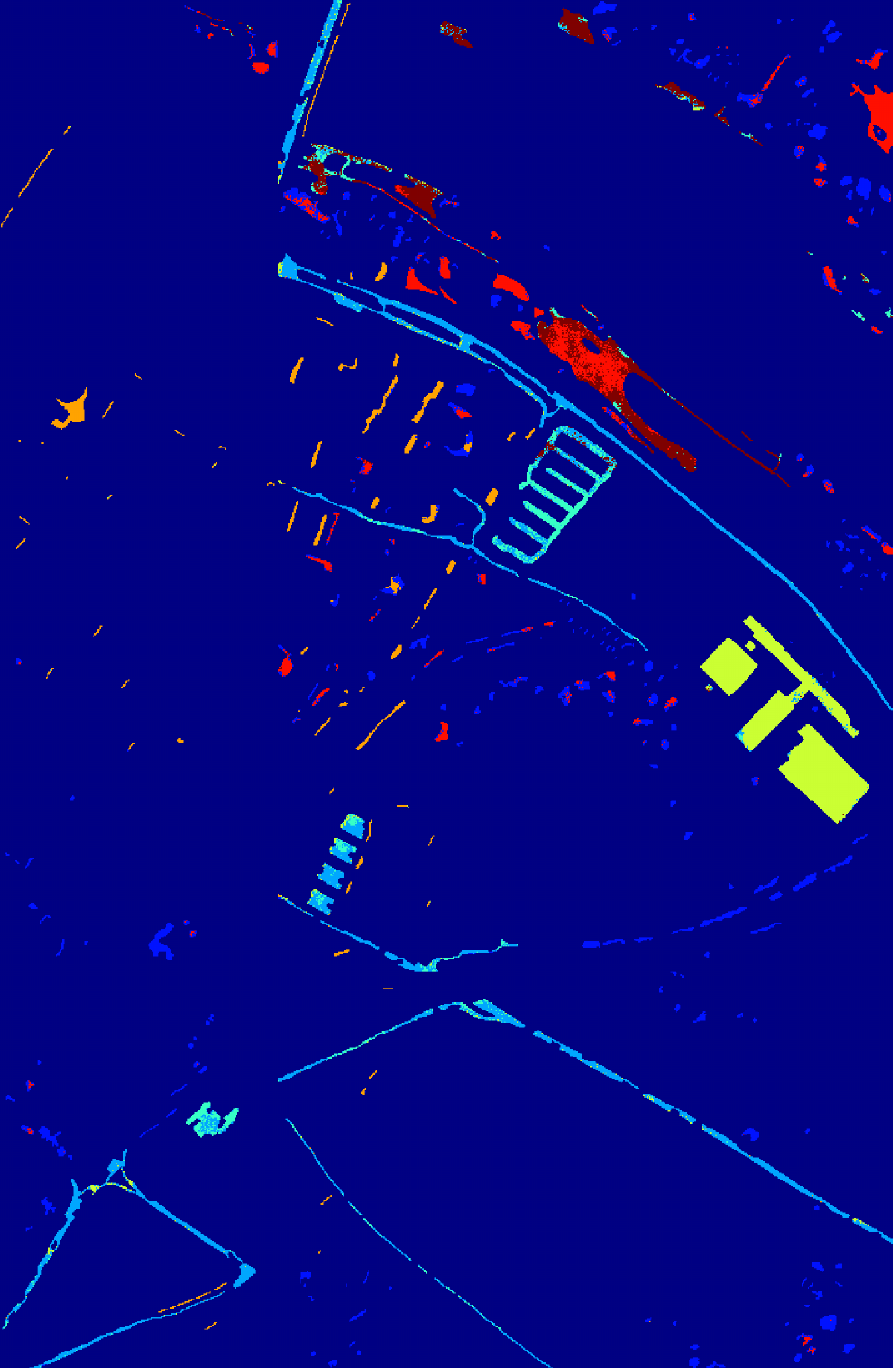}}   &
		\multicolumn{3}{c}{\epsfig{width=0.25\figurewidth,file=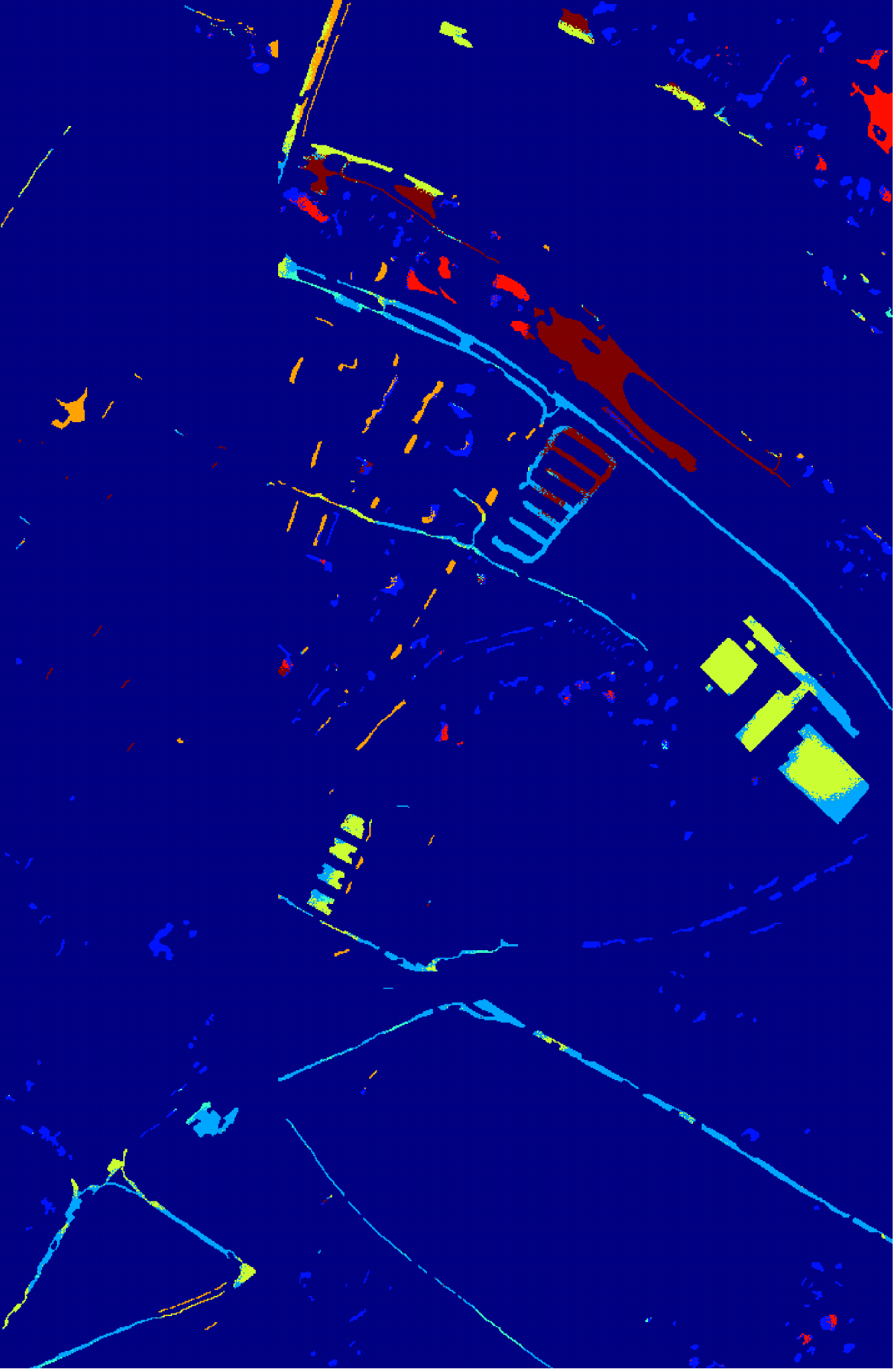}}   &
		\multicolumn{3}{c}{\epsfig{width=0.25\figurewidth,file=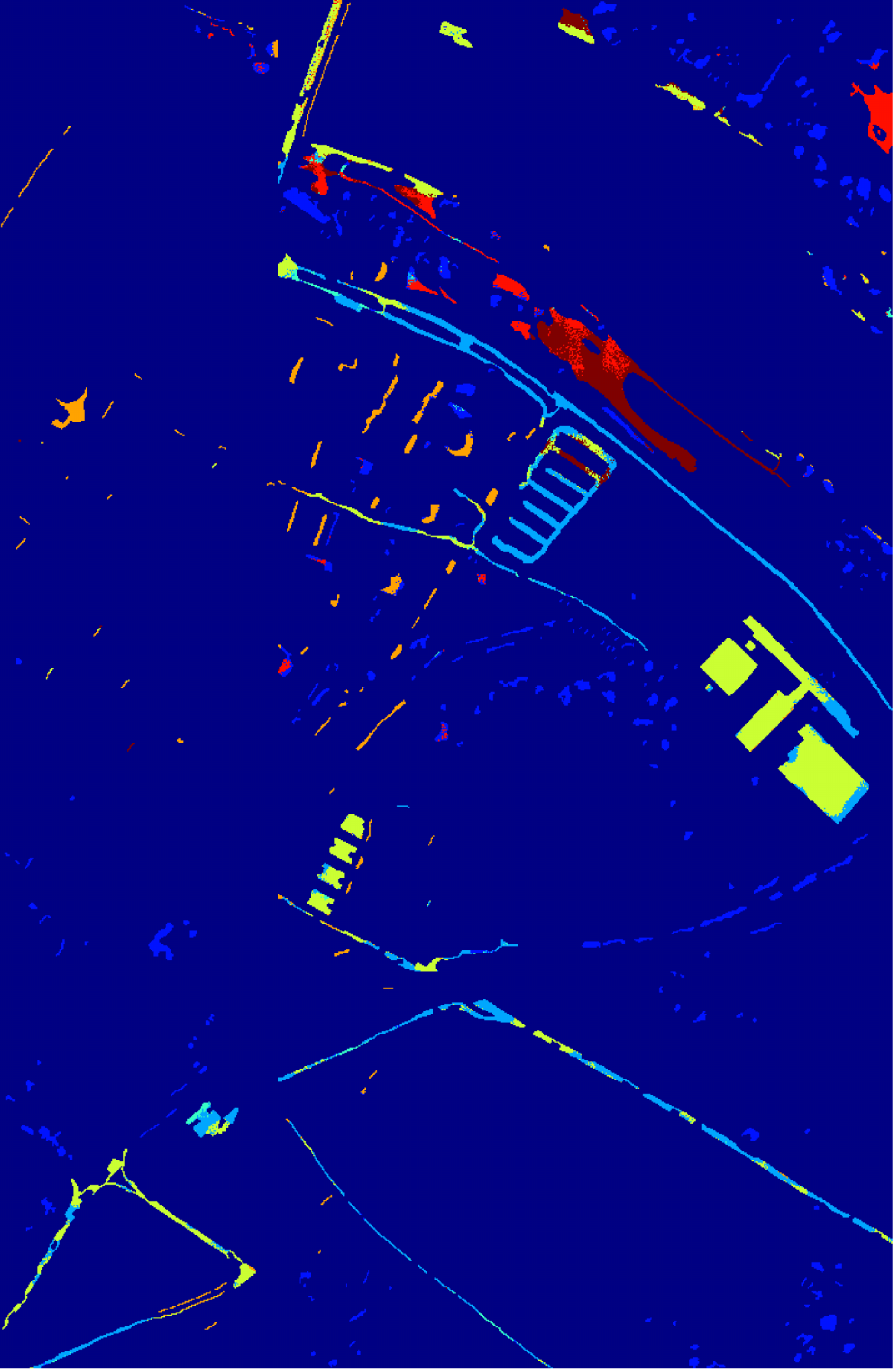}} &
		\multicolumn{3}{c}{\epsfig{width=0.25\figurewidth,file=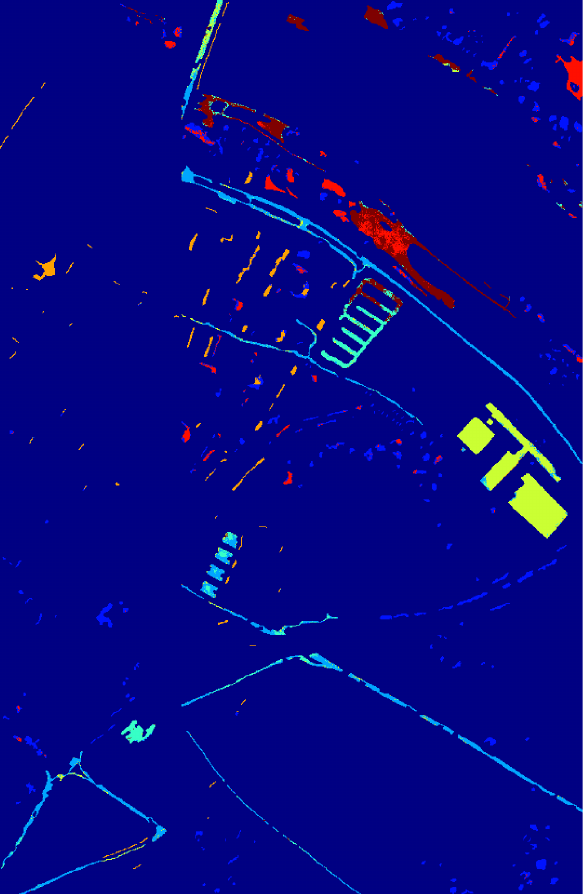}}   \\
		\multicolumn{3}{c}{(e)} & \multicolumn{3}{c}{(f)} &\multicolumn{3}{c}{(g)} &		\multicolumn{3}{c}{(h)} \\
	\end{tabular}
	\caption{\label{fig:Map_paviaU}
		Data visualization and classification maps for target scene Pavia Center data obtained with different methods including: (a) DAAN (65.62\%), (b) MRAN (69.22\%), (c) DSAN (78.94\%), (d) HTCNN (68.75\%), (e) PDEN (80.87\%), (f) LDSDG (71.02\%), (g) SagNet (69.90\%), (h)SDEnet (81.76\%).  }
	\vspace{-1.5em}
\end{figure}


\begin{figure*}[tp]
	\centering
	\begin{tabular}{ccccccccccccccc}
		\multicolumn{2}{c}{\epsfig{width=0.52\figurewidth,file=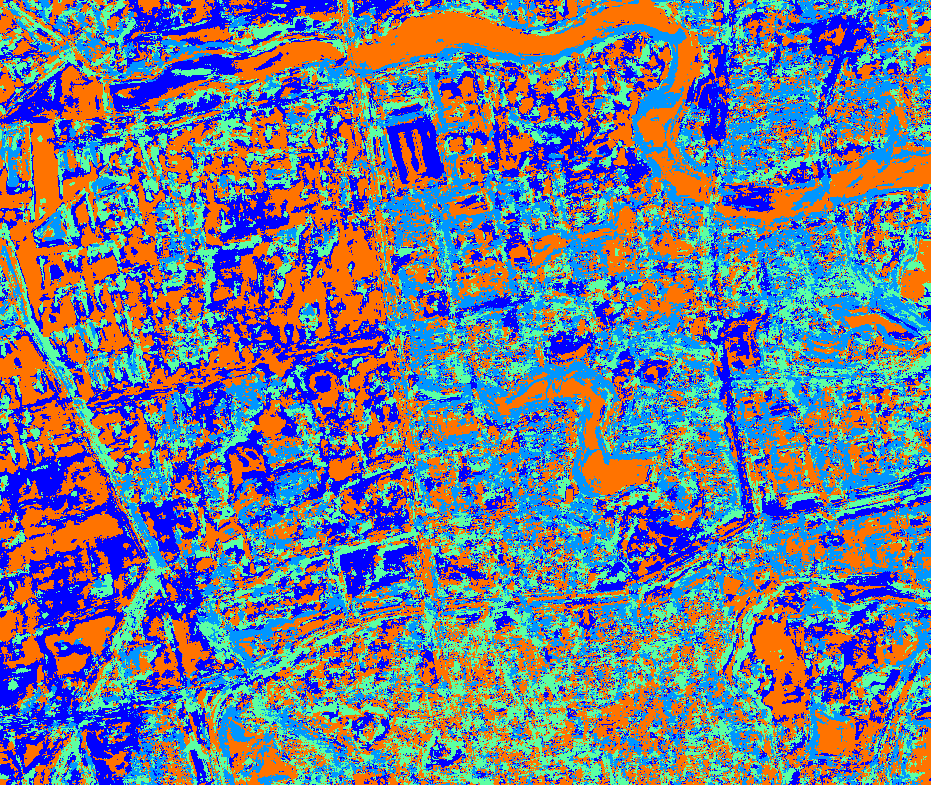}}     &
		\multicolumn{2}{c}{\epsfig{width=0.52\figurewidth,file=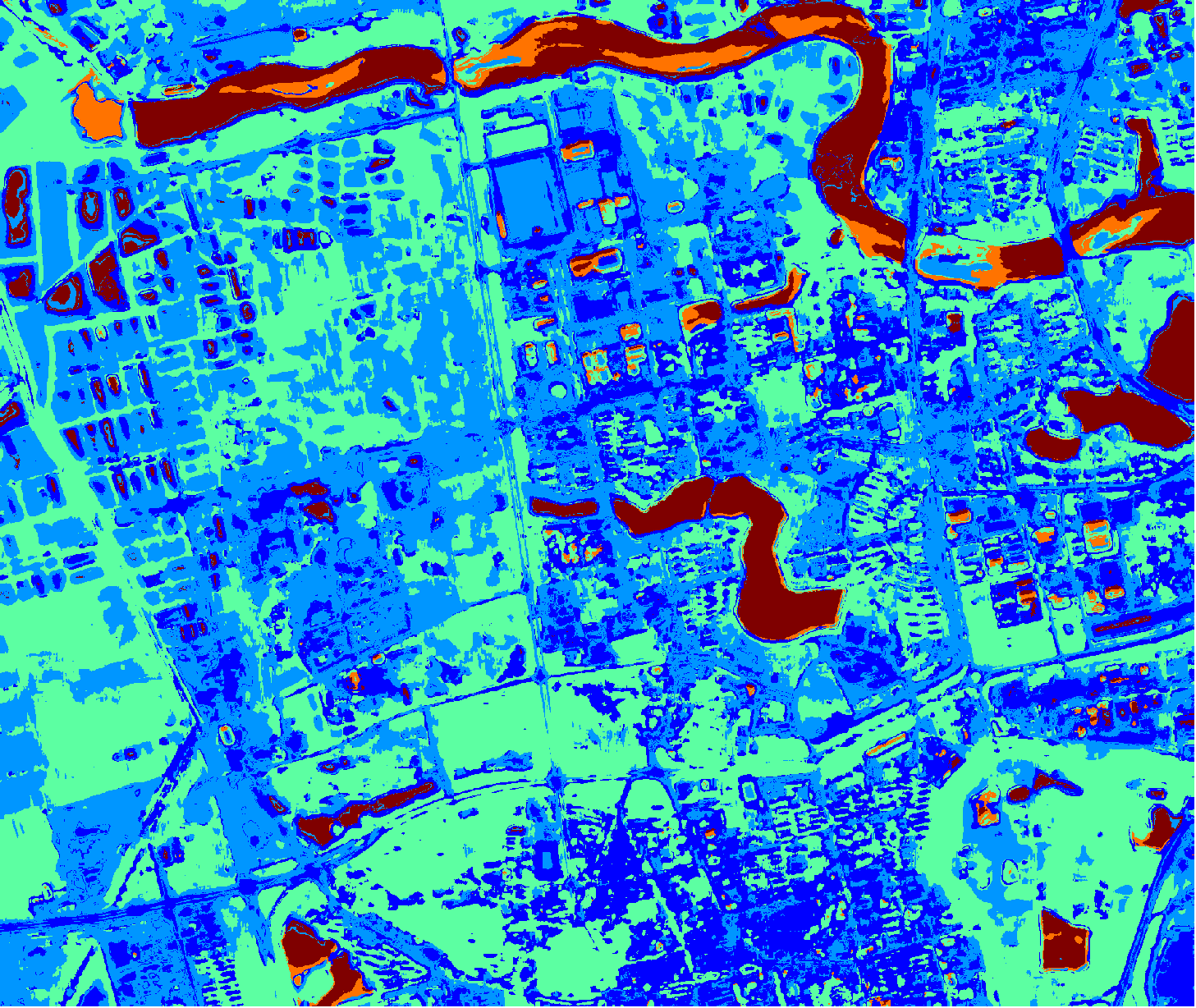}}  & 		\multicolumn{2}{c}{\epsfig{width=0.52\figurewidth,file=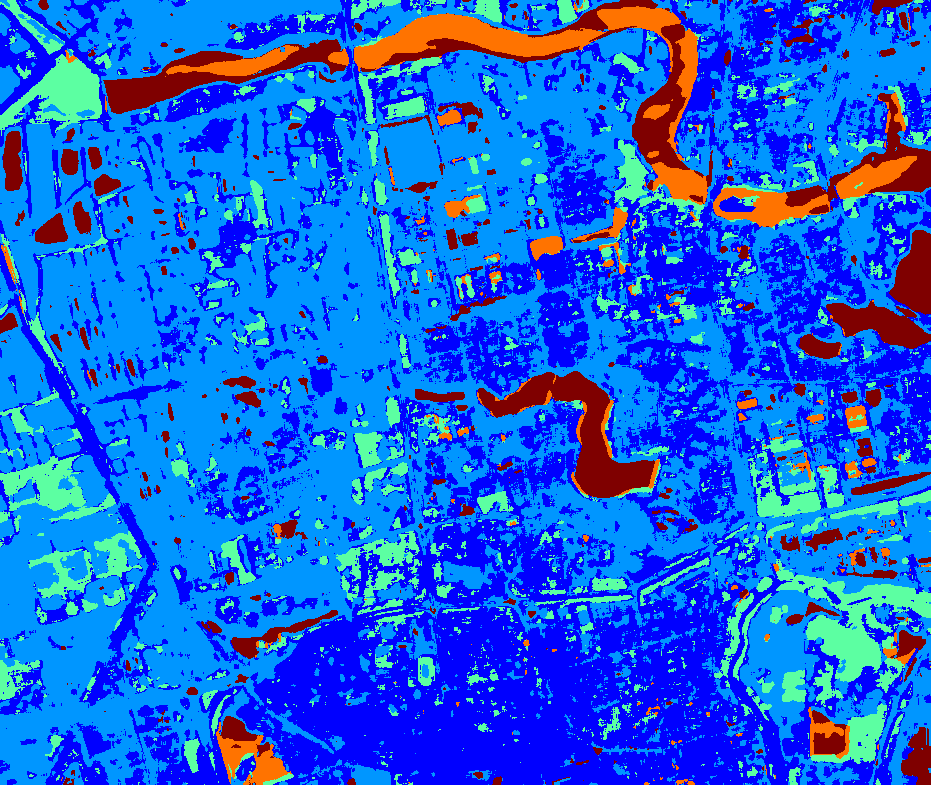}}    &		\multicolumn{2}{c}{\epsfig{width=0.52\figurewidth,file=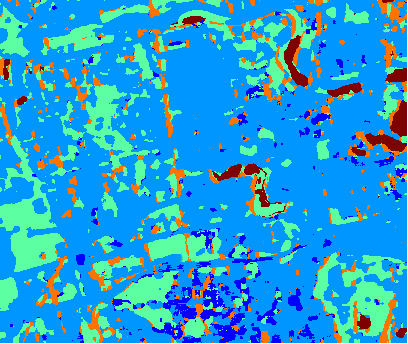}}   \\
		\multicolumn{2}{c}{(a)} & \multicolumn{2}{c}{(b)} & \multicolumn{2}{c}{(c)}  & \multicolumn{2}{c}{(d)}  \\
		\multicolumn{2}{c}{\epsfig{width=0.52\figurewidth,file=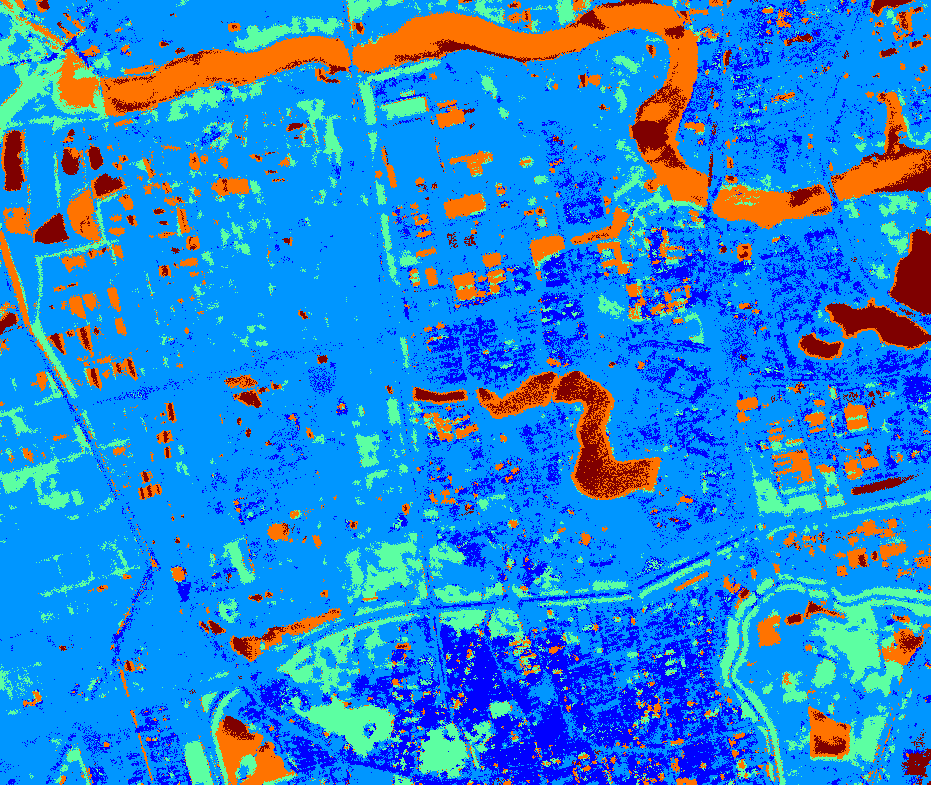}}   &
		\multicolumn{2}{c}{\epsfig{width=0.52\figurewidth,file=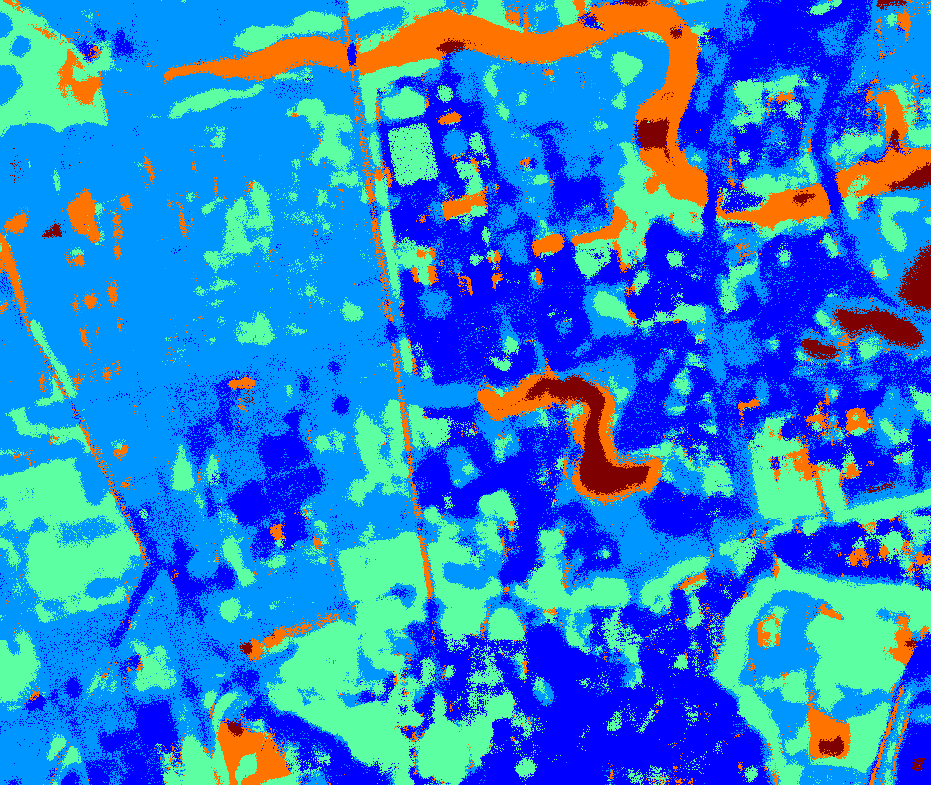}} &
		\multicolumn{2}{c}{\epsfig{width=0.52\figurewidth,file=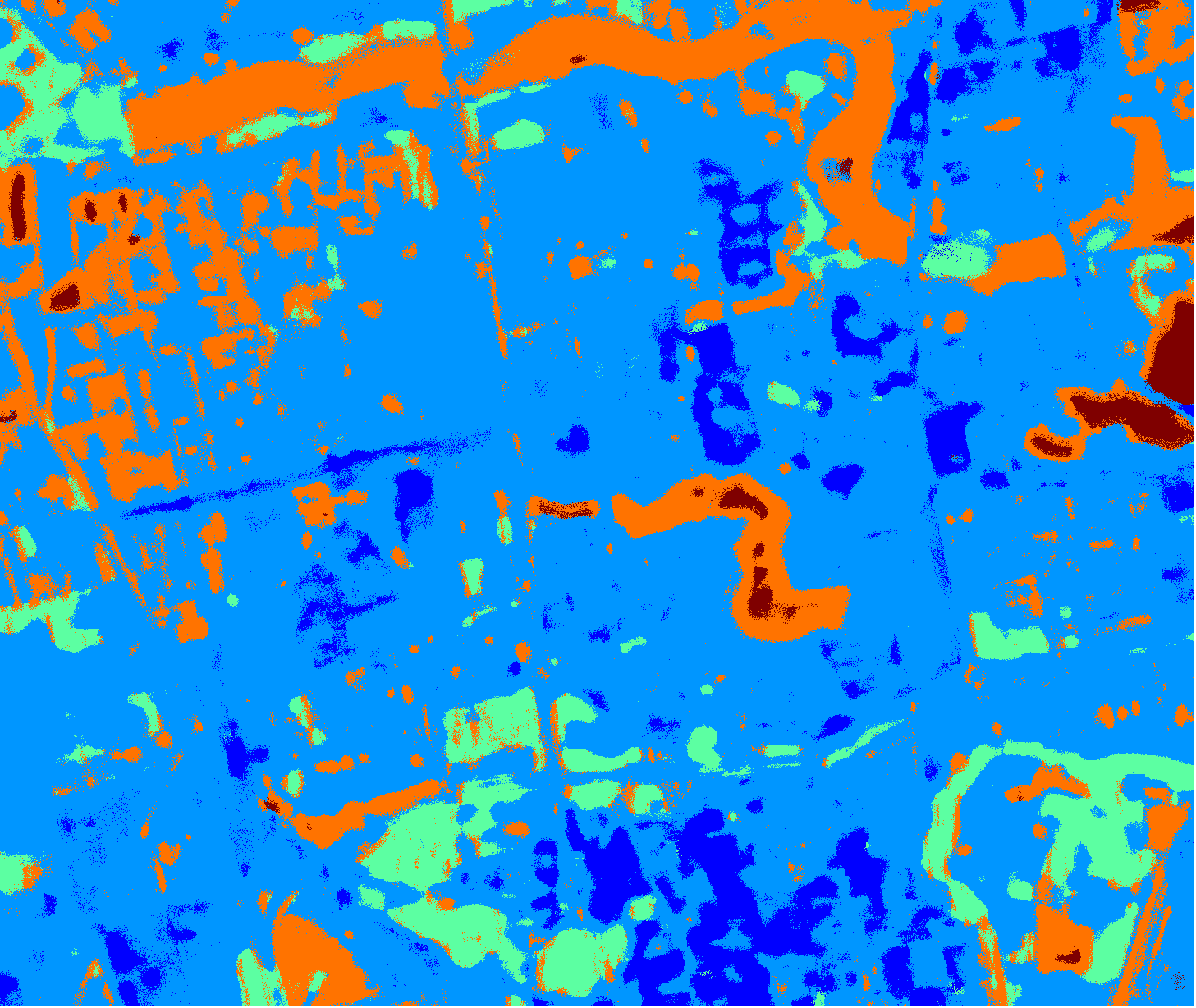}} &
		\multicolumn{2}{c}{\epsfig{width=0.52\figurewidth,file=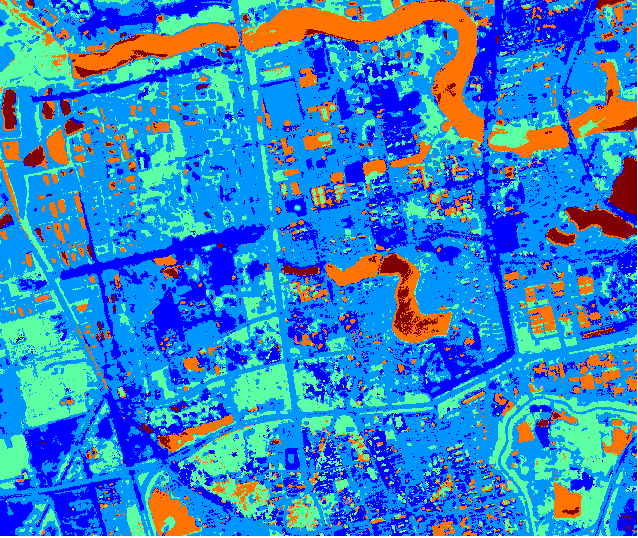}}  \\
		\multicolumn{2}{c}{(e)} &\multicolumn{2}{c}{(f)}& \multicolumn{2}{c}{(g)} & \multicolumn{2}{c}{(h)}\\
	\end{tabular}
	
	\caption{\label{fig:Map_GID}
		Data visualization and classification maps for target scene GID-wh data obtained with different methods including: (a) DAAN (68.06\%), (b) MRAN (67.49\%), (c) DSAN (73.69\%), (d) HTCNN (57.20\%), (e) PDEN (67.71\%), (f) LDSDG (76.48\%), (g) SagNet (61.64\%), (h)SDEnet (78.24\%).  }
\end{figure*}

\begin{table*}[tp]
	\vspace{-2em}
	\caption{\label{table:time}
		The execution time (in seconds) of one epoch training in different methods.}
	\begin{center}
		\begin{tabular}{|c|c|c|c|c|c|c|c|c|}
			\hline\hline
			Methods    & {DAAN \cite{yu2019transfer}}&  {MRAN \cite{ZHU2019214}} & {DSAN \cite{2020Deepsub}} & {HTCNN \cite{2019Heterogeneous}}  & {PDEN \cite{li2021progressive}}  & {LDSDG \cite{wang2021learning}}  & {SagNet \cite{nam2021reducing}}  & {SDEnet} \\ \hline
			Houston 2018 & 15.68 & 13.52 & 16.64 & 31.21 & 4.88  & 43.56  & 7.03   & 5.74   \\ \hline
			Pavia Center & 28.32 & 29.01 & 31.76 & 46.24 & 14.21 & 63.38 & 20.44  & 17.56  \\ \hline
			GID-wh       & 20.35 & 22.32 & 24.43 & 39.65 & 3.15  & 20.78  & 16.78  & 6.52   \\ \hline \hline
		\end{tabular}
	\end{center}
\vspace{-2em}
\end{table*}
\subsection{Parameter Tuning}

A parameter sensitivity analysis is conducted to evaluate the sensitivity of SDEnet on the three TDs. The base learning rate ${\eta}$, regularization parameters $\lambda$ and embedding feature dimension ${d_{se}}$ in semantic encoder, regarded as adjustable hyperparameters are selected from \{$1e-5$, $1e-4$, $1e-3$, $1e-2$, $1e-1$\}, \{$1e-3$, $1e-2$, $1e-1$, $1e+0$, $1e+1$\} and \{16, 32, 64, 128\}, respectively. 

In the gradient descent updates, the gradient of loss function is used to estimate hyperparameters of the model weight after being adjusted by the learning rate. Table \ref {tab:lr} provides classification results corresponding to different base learning rates in three data sets. The optimal base learning rate corresponding to three data sets is $1e-3$. The OA of all experimental datasets in regularization parameter $\lambda$ and embedding feature dimension ${d_{se}}$ are listed in Table \ref {tab:lambda} and Table \ref {tab:dse}. The optimal $\lambda$ is 1e-1 for all datsets, and ${d_{se}}$ is 64 for the Houston dataset and Pavia dataset, and 16 for the GID dataset.

SDEnet is implemented on the Pytorch platform. The input is set as patch size of 13$\times$13. Adaptive Moment Estimation(Adam) is used as the optimization scheme for generator and discriminator. The default value for ${\ell _2}$-norm regularization of all modules is set to 1e-4 for weight decay.

\subsection{Ablation Study}

The semantic encoder and morph encoder are the key components of generator, and contrastive learning and adversarial training are the main strategies for optimizing discriminator and generator. To assess the contribution of key components of SDEnet, ablation analyses are conducted by removing each component from the entire framework.

There are four variants in the ablation analyses, (1) ``SDEnet (no se)": the semantic encoder is deleted from generator, (2) ``SDEnet (no me)": the morph encoder is deleted, (3) ``SDEnet (no con)": the contrastive learning (${{\cal L}_{supcon}}$) is removed, (4) ``SDEnet (no adv)": the adversarial training (${{\cal L}_{adv}}$) is removed. As shown in Table \ref{table:Ablation}, it is obvious that the proposed SDEnet outperforms other variants and gains large improvements. The performance of either SDEnet without the semantic encoder (no se) or the morph encoder (no me) in the generator drops sharply, indicating that the two encoders play an important role in the effectiveness and reliability of ED. The most obvious is that in GID dataset, the OA of SDEnet (no me) is reduced by 13\%, because the GID has only four bands, and spectral randomization may distort the spectral information of ED. The template features with morphological knowledge are used to ensure that ED is not too far from SD. In addition, the classification performance of SDEnet (no con) drops from 1\% $\sim $ 9\% on all TDs, so the class-wise domain invariant representation can be learned only by constantly comparing the differences between SD and ED samples in the same and different classes during training. SDEnet (no adv) has the smallest drop in OA compared to other variants, but also optimizes the generator to improve the effectiveness of ED.

\subsection{Performance on Cross-Scene HSI Classification}
\label{sec:Classification Performance}
To evaluate the performance of SDEnet with only SD used for training, relevant algorithms including DAAN, MRAN, DSAN, HTCNN, PDEN, LDSDG and SagNet are used for comparison. The training samples is set as follows. DAAN, MRAN, DSAN and HTCNN regarded as DA methods, all data of SD with labels (80\% for training and 20\% for validation) and all TD data without labels are used for training. For DG methods, PDEN, LDSDG and SagNet, the selection of training samples is only SD with labels (80\% for training and 20\% for validation), where the patch size of LDSDG and SagNet is set to 32$\times$32 to fit the input size of Resnet18. In addition, the SD in Houston dataset is augmented by four times through random flip and random radiation noise (illumination), while the other two datasets are not augmented. The optimal base learning rate and regularization parameters of all comparison algorithms are selected from \{$1e-5$, $1e-4$, $1e-3$, $1e-2$, $1e-1$\} and \{$1e-3$, $1e-2$, $1e-1$, $1e+0$, $1e+1$, $1e+2$\}, respectively, and cross-validation is used to find the corresponding optimal parameters.

The following analyses are obtained from Tables \ref{tab:accuracy_Hou}-\ref{tab:accuracy_GID}.

\begin{itemize}
	\item The best performance of DA method on all TDs is DSAN. In the comparison of DG methods, PDEN performs well on Houston 2018 data and Pavia Center data, and LDSDG performs well on GID-wh data. In particular, DSAN provides 2\% improvement in OA over PDEN on Houston 2018 data, while PDEN and LDSDG are 2\% higher than DSAN on Pavia Center and GID-wh, respectively. This shows that DA method and DG method have their own advantages in different scenes, and TD is not necessarily used in the training process to achieve the best classification performance.
	
	\item SDEnet is improved by 4\% to 8\% over DSAN on all TDs. During the training process, DSAN directly accesses TD, and explicitly uses the domain alignment strategy to reduce the domain shift. However, ED is generated in SDEnet by the generator to make the domain shift change dynamically, and is used for discriminator to learn domain invariant representation. The improvement of SDEnet classification performance shows that the implicit and variable learning strategy is more effective than the domain alignment strategy in DA.
	
	\item Compared with the DG methods in computer vision, PDEN and LDSDG et al., SDEnet is increased by 1\% to 4\% on OA. This stems from the design of a more suitable generation method for HSI, semantic encoder and morph encoder, compared with those methods that only focus on spatial dimension changes.
	
	\item The proposed method is not only suitable for HSI, but also for MSI with only several bands. GID dataset is only four bands MSI, although the spectral dimension information is far less than HSI, it can be seen from the performance of SDEnet on GID (Table \ref{tab:accuracy_GID}) that the spatial-spectral generation strategy is still effective.
	
\end{itemize}

Classification maps are illustrated in Figs. \ref{fig:Map_Hou}-\ref{fig:Map_GID}. In Figs. \ref{fig:Map_Hou}-\ref{fig:Map_paviaU}, labeled pixels are displayed as ground truth and unlabeled pixels as backgrounds, and all pixels are predicted for comparison in Fig. \ref{fig:Map_GID}. In contrast, the proposed SDEnet obtains less noisy and more accurate results in some areas of the classification maps, such as the 2-nd class (Grass stressed) in Houston 2018 data and the 3-rd class (Brick) in Pavia Center data, where 3-rd class (Brick) in Pavia Center data is greatly improved compared to all comparison methods. It is obvious from Fig. \ref{fig:Map_GID} that the 4-th (River) and 5-th (Lake) in GID-wh data are better predicted.

To show computational complexity of different methods, the one epoch training time on all experimental data are listed in Table \ref{table:time}. All the experiments are carried out using Pytorch on an AMD EPYC 7542 32-Core Processor (48-GB RAM) powered with Nvidia GTX 3090 GPU with 24GB memory. It can be seen that the computational cost of SDEnet is much lower than that of other comparison methods except PDEN. This is due to the fact that only two layers of Conv2D-Relu-MaxPool2D blocks are used to learn domain invariant representation, which is much less complex than the DA and DG methods using VGG or Resnet as backbone. In addition, the double-branch encoder design does not bring additional computational cost to the model, and is lower in complexity than the Style-Complement module designed in LDSDG, which considers multiple potential style variations.

\section{Conclusions}
\label{sec:conclusions}
Single-source Domain Expansion Network (SDEnet), a domain generalization framework for cross-scene HSI classification, has been proposed. It can be generalized to target domain (TD) by using only source domain (SD) through generative adversarial learning. Specifically, the generator is designed based on the architecture of encoder-randomization-decoder. The semantic encoder uses spatial and spectral randomization and the morph encoder extracts template features, resulting in an extended domain (ED). In the discriminator, the supervised contrastive learning is employed to learn class-wise domain invariant representations. Furthermore, an adversarial training with supervised contrastive learning is designed to make ED  have a certain level of domain shift, so as to be dissimilar to SD. Comprehensive experiments on three datasets verify the effectiveness of the proposed SDEnet in domain extension. It offers the performance comparable to or even better than domain adaptation methods using TD data for model training.

\bibliographystyle{IEEEtran}
\bibliography{bibfile_zyx}

\end{document}